\documentclass[10pt,a4paper]{article}
\usepackage{hyinv_report}
\provisionalfalse\interimappsfalse  

\newcommand{\AblDeltaMPNNCathFour}{$+0.0082$}

\newcommand{\AblDeltaPZCathFour}{$+0.0016$}

\newcommand{\AblDeltaPZCplt}{$-0.0034$}

\newcommand{\AdvDivisor}{0.1}
\newcommand{\ApxEffArgmax}{99.73\%}

\newcommand{\ApxEffCritical}{0.002}
\newcommand{\ApxEffCuda}{12.9}
\newcommand{\ApxEffDZero}{0.30}
\newcommand{\ApxEffDivAgree}{0.001}

\newcommand{\ApxEffKLMean}{$5.2\times10^{-5}$}
\newcommand{\ApxEffKLP}{$7.5\times10^{-4}$}
\newcommand{\ApxEffKLSFT}{$4.2\times10^{-5}$}
\newcommand{\ApxEffLogitMax}{0.069}
\newcommand{\ApxEffLooShare}{1.01$\times$}
\newcommand{\ApxEffLooSubset}{1.16$\times$}
\newcommand{\ApxEffLooW}{1.24$\times$}
\newcommand{\ApxEffNumBackbones}{200}

\newcommand{\ApxEffTmax}{0.5}
\newcommand{\ApxEffTorch}{2.7.1}

\newcommand{\BaseDPLMSeqIdent}{99\%}
\newcommand{\BaseDivAllMPNNCathFour}{0.290}

\newcommand{\BaseDivDPLMCathFour}{0.000}

\newcommand{\BaseDivMPNNCathFour}{0.314}

\newcommand{\BaseDivPZCathFour}{0.315}

\newcommand{\BaseFrontDeltaCathFour}{$+0.0146$ $[+0.0058, +0.0242]$}
\newcommand{\BaseFrontDeltaCplt}{$+0.0187$ $[+0.0078, +0.0302]$}

\newcommand{\BaseGate}{0.003}
\newcommand{\BaseGateDevMax}{$0.0022$}
\newcommand{\BaseGateDevMin}{$0.0001$}

\newcommand{\BaseNBaselines}{ten}
\newcommand{\BaseNCells}{70}

\newcommand{\BaseNPanelsESM}{five}
\newcommand{\BaseNPos}{70}

\newcommand{\BaseNSig}{61}

\newcommand{\BaseNZero}{9}

\newcommand{\BaseZeroList}{ProteinMPNN (TS500, TS50); ProteinMPNN grouped (TS50); ProteinZero V2 (ESM1) (CATH4.3 test, TS500, TS50); ProteinZero V2 (ESM2) (CATH4.3 test, TS50); LigandMPNN (TS50)}
\newcommand{\BootB}{10{,}000}

\newcommand{\CandPerUpdate}{8{,}192}
\newcommand{\CkptGPUs}{56}

\newcommand{\CohortSize}{256}
\newcommand{\ConfDivSameT}{0.184}

\newcommand{\DecLayers}{3}
\newcommand{\DecodePasses}{16}

\newcommand{\DemoDenoFoldThresh}{0.9}

\newcommand{\DemoDenoMPNNGridHi}{0.53}
\newcommand{\DemoDenoMPNNGridLo}{0.47}
\newcommand{\DemoDenoN}{74}
\newcommand{\DemoDenoNovelThresh}{30\%}

\newcommand{\DemoDenoYield}{0.38}
\newcommand{\DemoDenoYieldMPNN}{0.05}
\newcommand{\DemoDenoYieldMPNNU}{0.47}
\newcommand{\DemoDenoYieldPZ}{0.11}
\newcommand{\DemoDenoYieldPZU}{0.45}
\newcommand{\DemoDenoYieldU}{0.68}
\newcommand{\DemoEnzN}{128}

\newcommand{\DemoFiltEntropy}{3.5}
\newcommand{\DemoFiltHomopolymer}{6}
\newcommand{\DemoFiltHydHi}{0.55}
\newcommand{\DemoFiltHydLo}{0.25}
\newcommand{\DemoLoopPool}{128}
\newcommand{\DemoLoopRMSDHi}{1.8\,\AA{}}
\newcommand{\DemoLoopRMSDLo}{0.5\,\AA{}}

\newcommand{\DemoLoopWinHi}{25}
\newcommand{\DemoLoopWinLo}{16}

\newcommand{\DemoPocketLen}{148}
\newcommand{\DemoPocketMeanPass}{43\%}
\newcommand{\DemoPocketN}{384}
\newcommand{\DemoPocketNBack}{twelve}
\newcommand{\DemoPocketNFixed}{15}
\newcommand{\DemoPocketPass}{373}

\newcommand{\DevD}{0.30}

\newcommand{\DevGainRL}{$+0.0133$}

\newcommand{\DevSadRL}{0.852}

\newcommand{\DevSadSFT}{0.839}
\newcommand{\DevSlope}{0.07}

\newcommand{\DivCathFour}{0.331}

\newcommand{\DivCostMax}{0.0011}

\newcommand{\DivDeficitCpltPZ}{0.015}

\newcommand{\DivMaxExcess}{0.016}

\newcommand{\DraftRounds}{8}

\newcommand{\EffAblationEnd}{50.7~ms}
\newcommand{\EffAblationEndRound}{about 51~ms}

\newcommand{\EffAblationStartSec}{2.6~s}

\newcommand{\EffAgreeSeqFull}{49.4\%}
\newcommand{\EffAgreeSiteFull}{98.7\%}
\newcommand{\EffBatch}{32}
\newcommand{\EffDivAgree}{0.001}
\newcommand{\EffExactIdentEager}{1{,}600 of 1{,}600}
\newcommand{\EffExactIdentGraph}{400 of 400}
\newcommand{\EffExactIdentPadded}{352 of 352}

\newcommand{\EffExactSpeedupLong}{8.7}

\newcommand{\EffFirstCallHi}{834~ms}
\newcommand{\EffFirstCallLo}{67.7~ms}
\newcommand{\EffGPU}{NVIDIA H20}
\newcommand{\EffLatencyLong}{50.6~ms}

\newcommand{\EffLenLong}{402}

\newcommand{\EffLibBatchRate}{5{,}430~sequences per second}
\newcommand{\EffLibBigBatch}{128}
\newcommand{\EffLibEightGPUTime}{0.253~s}
\newcommand{\EffLibLen}{250}
\newcommand{\EffLibMPNNTime}{142.0~s}
\newcommand{\EffLibN}{10{,}000}
\newcommand{\EffLibRefTime}{129.1~s}

\newcommand{\EffLibTime}{2.11~s}
\newcommand{\EffLooAllGeluLong}{1.68$\times$}
\newcommand{\EffLooFoldLong}{1.27$\times$}

\newcommand{\EffMemoryMPNNMiB}{3{,}164~MiB}
\newcommand{\EffMemoryMiB}{811~MiB}
\newcommand{\EffMemoryRefMiB}{4{,}323~MiB}
\newcommand{\EffNDesigns}{128}

\newcommand{\EffNDevTargets}{1{,}529}

\newcommand{\EffPasses}{16}
\newcommand{\EffPrecision}{BF16 operands with FP32 accumulation}
\newcommand{\EffQualityStep}{288}

\newcommand{\EffRepsHeadline}{11}
\newcommand{\EffRepsLib}{3}
\newcommand{\EffRepsSweep}{7}
\newcommand{\EffRoundsMax}{0.0074}
\newcommand{\EffRoundsMin}{0.0028}

\newcommand{\EffSadASub}{$-0.0008$}

\newcommand{\EffSadBSub}{$+0.0005$}

\newcommand{\EffSadCSub}{$+0.0001$}
\newcommand{\EffSadDelta}{$-0.0001$}

\newcommand{\EffSadInterval}{$[-0.0009, +0.0016]$}

\newcommand{\EffScaling}{7.7}

\newcommand{\EffSpeedupInt}{52}
\newcommand{\EffSpeedupLong}{52.1}

\newcommand{\EffSpeedupMPNNLong}{57}

\newcommand{\EffStabRange}{0.5\%}
\newcommand{\EffThroughput}{4{,}740~sequences per second}
\newcommand{\EffThroughputEightGPU}{39{,}527~sequences per second}
\newcommand{\EffTimingTr}{0.23}
\newcommand{\EffWarmup}{3}
\newcommand{\EncLayers}{4}
\newcommand{\EquivCoordDiff}{0.53~\AA}
\newcommand{\EquivMedianDScTM}{$9\times10^{-5}$}
\newcommand{\EquivNSeqs}{78}
\newcommand{\EquivSelfDiff}{0.36~\AA}
\newcommand{\FoldDiffusionSteps}{50}
\newcommand{\FoldLoops}{3}
\newcommand{\FoldSpeedup}{1.2 to 1.45 times}
\newcommand{\GPUsEarly}{32}
\newcommand{\GPUsFinal}{56}
\newcommand{\GPUsLate}{80}
\newcommand{\GPUsMid}{72}

\newcommand{\GainDraftSixCathFour}{$+0.0078$}
\newcommand{\GainDraftSixCplt}{$+0.0131$}

\newcommand{\GroupSize}{32}

\newcommand{\KHeld}{4}
\newcommand{\KLInit}{0.005}
\newcommand{\KLLate}{0.001}

\newcommand{\KRef}{2}

\newcommand{\MainAvgGainMPNN}{$+0.013$}

\newcommand{\MainAvgGainPZ}{$+0.008$}

\newcommand{\MainAvgScTMHyInv}{0.875}

\newcommand{\MainAvgScTMMPNN}{0.863}

\newcommand{\MainAvgScTMPZ}{0.867}

\newcommand{\MainGainMax}{$+0.062$}

\newcommand{\MainGainMin}{$+0.008$}

\newcommand{\MainNCols}{seven}

\newcommand{\MetDivAboveMPNNList}{CATH4.3 test and CATH4.2 test}
\newcommand{\MetDivAbovePZList}{CATH4.3 test and CATH4.2 test}

\newcommand{\MetSeed}{20260922}

\newcommand{\MinPValue}{0.1}
\newcommand{\NClustersCathFour}{103}

\newcommand{\NClustersCathTwo}{109}
\newcommand{\NClustersCplt}{121}
\newcommand{\NClustersDev}{91}
\newcommand{\NClustersTSLarge}{500}
\newcommand{\NClustersTSSmall}{50}

\newcommand{\NTargetsCathFour}{1{,}882}
\newcommand{\NTargetsCathTwo}{1{,}120}
\newcommand{\NTargetsCplt}{121}

\newcommand{\NTargetsDev}{1{,}529}
\newcommand{\NTargetsTSLarge}{500}
\newcommand{\NTargetsTSSmall}{50}

\newcommand{\Neighbours}{48}
\newcommand{\NoiseFoldOnlyMax}{0.0002}

\newcommand{\NoisePairedDiffs}{$-0.0008$ and $+0.0016$}
\newcommand{\NoiseSingleEval}{$\pm0.0015$}

\newcommand{\ObsImbalance}{1.04 to 1.08}
\newcommand{\ParamsAdj}{1.96-million-parameter}
\newcommand{\ParamsExact}{1{,}957{,}573}
\newcommand{\ParamsM}{1.96~million}
\newcommand{\PlanImbalance}{1.00 to 1.02}

\newcommand{\PoolExp}{16{,}839}

\newcommand{\PoolMeanLen}{254}
\newcommand{\PoolPred}{33{,}161}

\newcommand{\PoolSize}{50{,}000}

\newcommand{\RLBetas}{(0.9, 0.95)}

\newcommand{\RLLR}{$2\times10^{-5}$}
\newcommand{\RLPriorStep}{288}
\newcommand{\RLStep}{640}
\newcommand{\RLTemp}{0.3}
\newcommand{\RLTieWindow}{0.002}

\newcommand{\RecentPDBCutoff}{2025}
\newcommand{\RecentPDBEntries}{111}
\newcommand{\RecentPDBLenMax}{978}
\newcommand{\RecentPDBLenMin}{42}
\newcommand{\RecentPDBResidues}{26{,}766}
\newcommand{\ReplayErrMax}{$6.3\times10^{-5}$}

\newcommand{\RewriteDivSameT}{0.335}
\newcommand{\RewriteRounds}{8}

\newcommand{\RunUpdates}{656}
\newcommand{\SadB}{1{,}000}

\newcommand{\SadRefCathFour}{0.32}
\newcommand{\SadRefCplt}{0.30}
\newcommand{\SadRefDev}{0.30}

\newcommand{\SelGap}{$-0.0004$}

\newcommand{\SftBetas}{(0.9, 0.98)}

\newcommand{\SftExp}{76{,}787}
\newcommand{\SftLabelSmoothing}{0.1}
\newcommand{\SftMaxLen}{512}

\newcommand{\SftPred}{551{,}158}

\newcommand{\SftSeqs}{627{,}945}

\newcommand{\SftUpdates}{85{,}000}

\newcommand{\StepSecEarly}{275}
\newcommand{\StepSecFinal}{161}
\newcommand{\StepSecLate}{113}
\newcommand{\StepSecMid}{127}

\newcommand{\TdraftAltVal}{0.3}
\newcommand{\TdraftVal}{0.6}

\newcommand{\TopPValue}{0.9}

\newcommand{\TrewriteVal}{0.34}

\newcommand{\UpdFirstFinal}{529}

\newcommand{\UpdLastEarly}{320}

\newcommand{\UpdLastLate}{528}
\newcommand{\UpdLastMid}{405}

\newcommand{\Width}{128}
\renewcommand{\today}{2026-10-14}
\hypersetup{pdftitle={HyInv Technical Report},
 pdfauthor={Hyra Team},
 pdfsubject={HyInv, an inverse-folding model trained with structure-reward reinforcement learning},
 pdfkeywords={HyInv, inverse folding, protein design, reinforcement learning, sampling, diversity}}
\title{HyInv Technical Report}
\author{Hyra Team}
\begin{document}
\maketitle
\ifprovisional\draftline\fi

\begin{abstract}
Inverse folding designs amino-acid sequences for a given protein backbone, and screening needs many candidates that are both diverse and well folded.
We present \HyInv, a \ParamsAdj{} backbone-conditioned graph network with an encoder--decoder architecture, trained by Hyra 2.0, the research agent developed by our team: pretrained with random-order autoregressive sequence modeling, followed by structure-reward reinforcement learning. At inference, a grouped draft-and-rewrite sampler that uses separate draft and rewrite temperatures improves refolding TM-score at matched diversity.
An optimized implementation with BF16 numerics designs \EffNDesigns\ sequences for a \EffLenLong-residue backbone in \EffLatencyLong\ on a single \EffGPU{} GPU, about \EffSpeedupInt$\times$ faster than the reference sampler, and an exact FP32 path that returns the reference sequences is \EffExactSpeedupLong$\times$ faster.
Across five benchmark datasets and seven dataset--predictor combinations, \HyInv\ reaches a higher average scTM than all \BaseNBaselines\ external baselines at comparable or higher sequence diversity, with large margins over ProteinMPNN and ProteinZero V2.
The same default setting supports de novo backbones, pocket-constrained libraries, enzyme redesign with the catalytic site kept, and loop redesign. We have released our code and model at \url{https://github.com/Tencent-Hunyuan/Hyra-results/tree/main/AI4Science/hyinv}.

\end{abstract}

\begin{figure}[t!]
\centering\includegraphics[width=\linewidth]{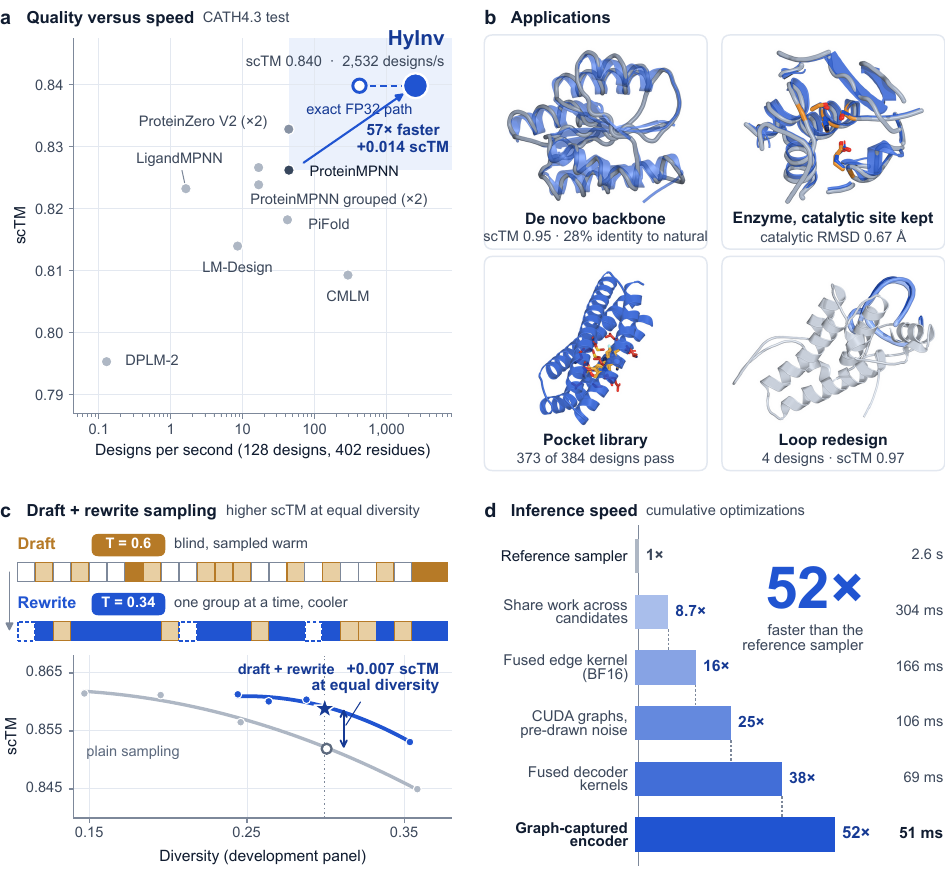}
\caption{\textbf{\HyInv\ designs sequences that fold like the input backbone, and does so fast.}
\textbf{a,} scTM against design speed on CATH4.3 test (ESMFold2) for \HyInv\ and \BaseNBaselines\ published baselines; speed is the time to design \EffNDesigns\ sequences for a \EffLenLong-residue backbone on one \EffGPU\ (logarithmic axis). scTM is measured on sequences from the reference sampler, which the exact FP32 path (open circle) reproduces; the BF16 path (filled) is faster still (Section~\ref{sec:efficiency}).
\textbf{b,} The same default setting applied to four single-chain design tasks (ribbons: designs in blue, input backbones in gray; the highest-scTM designs sare shown; pocket: the backbone shown, mean pass rate over \DemoPocketNBack\ backbones, \DemoPocketMeanPass).
\textbf{c,} Separate draft and rewrite temperatures give a higher scTM than plain sampling at equal diversity (development panel; star: \HyInv).
\textbf{d,} Cumulative inference optimizations reduce the time for \EffNDesigns\ designs from \EffAblationStartSec\ to \EffAblationEndRound.}
\label{fig:teaser}
\end{figure}

\section{Introduction}
Inverse folding designs amino-acid sequences that fold into a given backbone, and it underlies scaffold design, enzyme engineering and the construction of sequence libraries for screening.
Screening needs many diverse candidates rather than one best sequence, so a model is useful in proportion to the diversity it offers at a given structural quality.
ProteinMPNN made inverse folding routine~\citep{dauparas2022}, language-model-assisted and parallel decoders extended it~\citep{hsu2022,gao2023,zheng2023,wang2024dplm2}, and ProteinZero showed the value of reinforcement learning (RL) against a structure predictor~\citep{wang2025proteinzero}.
The structural consistency of designed sequences can be evaluated using scTM, the TM-score~\citep{zhang2004scoring} between their predicted structures and the input backbones. We use ESMFold2~\citep{esmfold2} as the primary structure predictor.
Evaluation at a single sampling temperature can obscure the trade-off between structural consistency and sequence diversity, so we compare models at comparable diversity.

We present \HyInv, a \ParamsAdj{} message-passing graph network~\citep{ingraham2019,dauparas2022} with an encoder--decoder architecture, trained by Hyra 2.0, the research agent developed by our team, in two stages: pretraining by maximizing sequence likelihood under a random residue decoding order on natural and predicted structures, and subsequent RL~\citep{rloo} primarily rewarded by predicted structural self-consistency on a pool of \PoolSize{} backbones filtered against the benchmark datasets with a homology mask.
At inference time, we introduce a draft-and-rewrite sampler that generates an initial sequence at a higher temperature and then refines it through groupwise resampling at a lower temperature, leading to a higher scTM while preserving diversity. An optimized implementation makes sampling fast enough to screen libraries in seconds.

Figure~\ref{fig:teaser} summarizes the results.
Averaged over the benchmark panels and structure predictors, \HyInv\ has a higher scTM than each of the \BaseNBaselines\ external baselines, at comparable or higher sequence diversity; the average margin is \MainAvgGainMPNN{} over ProteinMPNN and \MainAvgGainPZ{} over ProteinZero V2 (Figure~\ref{fig:teaser}a, Section~\ref{sec:results}).
Separate draft and rewrite temperatures raise scTM at equal diversity (Figure~\ref{fig:teaser}c).
An optimized implementation designs \EffNDesigns\ sequences for a \EffLenLong-residue backbone in \EffLatencyLong, \EffSpeedupLong$\times$ faster than the reference sampler (Figure~\ref{fig:teaser}d), and the same default setting serves four single-chain design tasks (Figure~\ref{fig:teaser}b).

Our contributions are:
\begin{itemize}\setlength\itemsep{1pt}
\item \textbf{A compact inverse-folding model with structure-based RL.} We train a \ParamsAdj{} graph network using random-order autoregressive pretraining and structure-reward RL (Section~\ref{sec:model}).
\item \textbf{Test-time scaling via draft-and-rewrite decoding.} We develop a two-temperature sampler that separates initial sequence construction from groupwise revision, which gives a higher scTM at equal diversity (Section~\ref{sec:sampling}).
\item \textbf{Inference optimization.} We implement batched inference across sequences that enables \EffSpeedupInt$\times$ faster sampling with BF16 numerics (Section~\ref{sec:efficiency}).
\item \textbf{Comprehensive benchmark evaluation and computational design case studies.} We report a five-panel comparative evaluation in which \HyInv\ achieves the highest scTM at a comparable or higher diversity relative to \BaseNBaselines\ baselines, and study four applications spanning practical tasks (Sections~\ref{sec:results}--\ref{sec:applications}).
\end{itemize}

\section{HyInv}\label{sec:model}
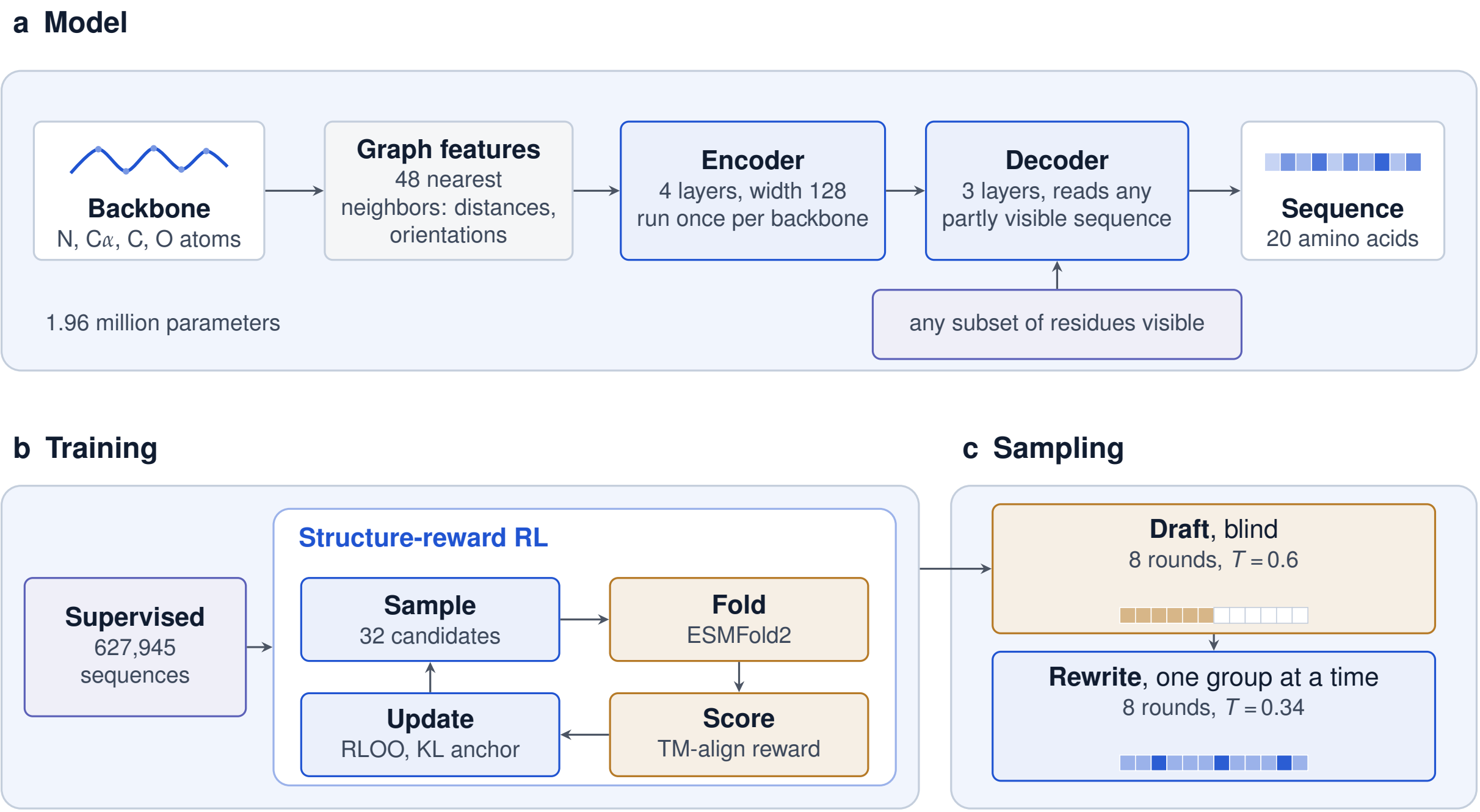
\begin{figure}[tbp]
\centering
\begin{adjustbox}{max width=\linewidth}
\begin{tikzpicture}[x=1cm,y=1cm,
  ttl/.style={font=\sffamily\bfseries\fontsize{8}{9.5}\selectfont},
  sub/.style={font=\sffamily\fontsize{7}{8.5}\selectfont,text=hymuted},
  lane/.style={font=\sffamily\bfseries\fontsize{9}{10.5}\selectfont,text=hyink,anchor=base west},
  nd/.style={hy/base,minimum height=0pt,inner xsep=2pt,inner ysep=2.5pt},
  gp/.style={hy/group,rounded corners=6pt},
  cell/.style={draw=white,line width=0.3pt}]
  \node[lane] at (0,0.42) {a\enspace Model};
  \begin{scope}[on background layer]\node[gp,fit={(0,0)(16,-3.25)},inner sep=0pt] (ga) {};\end{scope}
  \node[nd,hy/box,minimum width=25mm,minimum height=15mm,text depth=0pt] (a1) at (1.6,-1.3) {\rule{0pt}{5.6mm}\\ \textbf{Backbone}\\[-0.5pt]\hysub N, C$\alpha$, C, O atoms};
  \node[nd,hy/norm,minimum width=26mm,minimum height=15mm] (a2) at (4.85,-1.3) {\textbf{Graph features}\\[-0.5pt]\hysub \Neighbours{} nearest\\[-1pt]\hysub neighbors: distances,\\[-1pt]\hysub orientations};
  \node[nd,hy/attn,minimum width=26mm,minimum height=15mm] (a3) at (8.15,-1.3) {\textbf{Encoder}\\[-0.5pt]\hysub \EncLayers{} layers, width \Width\\[-1pt]\hysub run once per backbone};
  \node[nd,hy/attn,minimum width=26mm,minimum height=15mm] (a4) at (11.45,-1.3) {\textbf{Decoder}\\[-0.5pt]\hysub \DecLayers{} layers, reads any\\[-1pt]\hysub partly visible sequence};
  \node[nd,hy/box,minimum width=22mm,minimum height=15mm] (a5) at (14.55,-1.3) {\rule{0pt}{5.6mm}\\ \textbf{Sequence}\\[-0.5pt]\hysub 20 amino acids};
  \foreach \p/\q in {a1/a2,a2/a3,a3/a4,a4/a5} \draw[hy/arrow] (\p) -- (\q);
  \begin{scope}[shift={($(a1.north)+(0,-0.45)$)}]
    \draw[hyblue,line width=1.0pt,smooth] plot coordinates {(-0.85,-0.12) (-0.55,0.14) (-0.25,-0.1) (0.05,0.15) (0.35,-0.08) (0.62,0.12) (0.85,-0.05)};
    \foreach \x/\y in {-0.55/0.14,-0.25/-0.1,0.05/0.15,0.35/-0.08,0.62/0.12} \fill[hyblue!60!white] (\x,\y) circle (0.035);
  \end{scope}
  \begin{scope}[shift={($(a5.north)+(0,-0.45)$)}]
    \foreach \i/\c in {0/25,1/60,2/40,3/80,4/30,5/65,6/45,7/90,8/35,9/70} \draw[cell,fill=hyblue!\c] (\i*0.17-0.85,-0.1) rectangle ++(0.17,0.2);
  \end{scope}
  \node[nd,hy/emb,minimum width=40mm,minimum height=7.5mm] (av) at (11.45,-2.75) {\hysub any subset of residues visible};
  \draw[hy/arrow] (av.north) -- (a4.south);
  \node[sub,anchor=west] at (0.35,-2.75) {\ParamsM{} parameters};
  \node[lane] at (0,-4.2) {b\enspace Training};
  \begin{scope}[on background layer]\node[gp,fit={(0,-4.5)(9.95,-8.0)},inner sep=0pt] (gb) {};\end{scope}
  \node[nd,hy/emb,minimum width=24mm,minimum height=15mm] (b0) at (1.45,-6.25) {\textbf{Supervised}\\[-0.5pt]\hysub \SftSeqs{}\\[-1pt]\hysub sequences};
  \begin{scope}[on background layer]\node[hy/group,fill=white,draw=hyblue!45,fit={(2.95,-4.75)(9.7,-7.75)},inner sep=0pt] (gr) {};\end{scope}
  \node[ttl,text=hyblue,anchor=north west] at (3.1,-4.83) {Structure-reward RL};
  \node[nd,hy/attn,minimum width=28mm,minimum height=9mm] (r1) at (4.65,-5.95) {\textbf{Sample}\\[-0.5pt]\hysub \GroupSize{} candidates};
  \node[nd,hy/mlp,minimum width=28mm,minimum height=9mm] (r2) at (8.0,-5.95) {\textbf{Fold}\\[-0.5pt]\hysub ESMFold2};
  \node[nd,hy/mlp,minimum width=28mm,minimum height=9mm] (r3) at (8.0,-7.2) {\textbf{Score}\\[-0.5pt]\hysub TM-align reward};
  \node[nd,hy/attn,minimum width=28mm,minimum height=9mm] (r4) at (4.65,-7.2) {\textbf{Update}\\[-0.5pt]\hysub RLOO, KL anchor};
  \draw[hy/arrow] (r1) -- (r2);
  \draw[hy/arrow] (r2) -- (r3);
  \draw[hy/arrow] (r3) -- (r4);
  \draw[hy/arrow] (r4) -- (r1);
  \draw[hy/arrow] (b0.east) -- (b0.east -| gr.west);
  \node[lane] at (10.3,-4.2) {c\enspace Sampling};
  \begin{scope}[on background layer]\node[gp,fit={(10.3,-4.5)(16,-8.0)},inner sep=0pt] (gc) {};\end{scope}
  \node[nd,hy/base,fill=hywarm!12,draw=hywarm,minimum width=48mm,minimum height=14mm] (c1) at (13.15,-5.4) {\textbf{Draft}, blind\\[-0.5pt]\hysub \DraftRounds{} rounds, \textit{T}\,=\,\TdraftVal\\[2.2mm]\rule{0pt}{0.1mm}};
  \node[nd,hy/attn,minimum width=48mm,minimum height=14mm] (c2) at (13.15,-7.0) {\textbf{Rewrite}, one group at a time\\[-0.5pt]\hysub \RewriteRounds{} rounds, \textit{T}\,=\,\TrewriteVal\\[2.2mm]\rule{0pt}{0.1mm}};
  \draw[hy/arrow] (c1) -- (c2);
  \draw[hy/arrow] (gb.east |- c1) -- (c1.west -| gc.west) -- (c1.west);
  \foreach \i/\f in {0/1,1/1,2/1,3/1,4/1,5/1,6/0,7/0,8/0,9/0,10/0,11/0} {%
    \ifnum\f=1 \draw[cell,fill=hywarm!55] ($(c1.south)+(\i*0.17-1.02,0.12)$) rectangle ++(0.17,0.17);%
    \else\draw[cell,draw=hyline,fill=white] ($(c1.south)+(\i*0.17-1.02,0.12)$) rectangle ++(0.17,0.17);\fi}
  \foreach \i/\c in {0/45,1/45,2/95,3/45,4/45,5/45,6/95,7/45,8/45,9/45,10/95,11/45} \draw[cell,fill=hyblue!\c] ($(c2.south)+(\i*0.17-1.02,0.12)$) rectangle ++(0.17,0.17);
\end{tikzpicture}
\end{adjustbox}
\caption{\textbf{Framework of \HyInv.}
\textbf{a,} The graph model: backbone features, an encoder that runs once per backbone, and a decoder that predicts amino acid distributions at masked positions using the observed sequence context.
\textbf{b,} Structure-reward RL: sample candidates for a backbone, fold them with ESMFold2, score them against the input backbone with TM-align, and update the policy with a leave-one-out advantage and a KL penalty to the supervised checkpoint.}
\label{fig:framework}
\end{figure}

\HyInv\ is a small graph network that maps a backbone to sequences, trained first by supervision on natural structures and then by reinforcement learning against a structure predictor (Figure~\ref{fig:framework}).

\subsection{Architecture}\label{sec:arch}
\HyInv\ follows the message-passing encoder--decoder of ProteinMPNN~\citep{dauparas2022,ingraham2019} and has \ParamsExact{} parameters.
The input is the N, C$\alpha$, C and O coordinates of $L$ residues.
Each residue is connected to its \Neighbours{} nearest neighbors, and edges carry rotation- and translation-invariant distance features.
The encoder has \EncLayers{} layers of width \Width{} and runs once per backbone; the decoder has \DecLayers{} layers of the same width.
At inference the model needs neither a protein language model, a structure predictor nor an alignment.

The decoder takes a partially masked sequence as input and predicts amino acid distributions at masked positions, conditioned on the backbone and the observed sequence context.
During sampling, the observed context consists of fixed or previously sampled amino acid identities supplied to the decoder at the current step.
With a random order this is an ordinary autoregressive model, which is the objective of both training stages, and any partially masked sequence is a legal input, which is what the sampler of Section~\ref{sec:sampling} uses.
Fixed residues retain their specified amino acid identities and remain unmasked throughout sampling.

\subsection{Supervised stage}
The network is trained from scratch on \SftSeqs{} sequences with structures, \SftExp{} experimental and \SftPred{} predicted, to maximize the likelihood of the native sequence under a random residue order, for \SftUpdates{} updates.
The checkpoint is chosen by validation scTM and serves as the starting point and the fixed reference of the RL stage (Appendix~\ref{app:training}).

\subsection{Structure-reward RL}
The RL stage rewards sequences whose predicted structure matches the input backbone.
For a backbone $\bb$ and a candidate sequence $\aa$ the reward is
\begin{equation}
\rew(\bb,\aa)=0.95\,\mathrm{scTM}(\bb,\aa)+0.05\,s(\bb,\aa),
\end{equation}
where scTM comes from ESMFold2~\citep{esmfold2} and $s\in[0,1]$ is a small stability-proxy term computed with fixed ProteinMPNN weights (Appendix~\ref{app:training}).
Native sequence recovery is not rewarded.
Boltz2 and Protenix2 are not part of the reward; they serve as independent predictors in the evaluation (Section~\ref{sec:results}).

Each update samples \GroupSize{} candidates for each of \CohortSize{} backbones from the policy and scores them with ESMFold2.
The advantage of a candidate is its reward minus the mean reward of the other candidates of the same backbone~\citep{rloo}, with a KL penalty to the supervised checkpoint (Appendix~\ref{app:training}).
The \PoolSize{} training backbones were filtered against the benchmark panels with a same-source homology mask.
The final checkpoint is update \RLStep{}; on the development panel and with plain sampling, its scTM at equal diversity is higher than that of the supervised checkpoint by \DevGainRL{} (Section~\ref{sec:ablation}).

\section{Sampling}\label{sec:sampling}
\HyInv\ draws designs with a draft-and-rewrite sampler in which the draft and the rewrite use different temperatures, which gives a higher scTM at equal diversity.

\subsection{Draft and rewrite}\label{sec:rewrite}
\begin{figure}[tbp]
\centering
\tikzset{
  hy/cell/mask/.style={fill=white,draw=hyline,line width=0.3pt},
  hy/cell/draft/.style={fill=hywarm!38,draw=white,line width=0.3pt},
  hy/cell/drafting/.style={fill=hywarm,draw=white,line width=0.3pt},
  hy/cell/rewritten/.style={fill=hyblue!38,draw=white,line width=0.3pt},
  hy/cell/rewriting/.style={fill=hyblue,draw=white,line width=0.3pt}}
\providecommand{\hystrip}[4]{
  \begin{scope}[shift={(#1,#2)}]
    \foreach \g [count=\i from 0] in {3,0,5,2,7,1,4,6,0,5,2,7,3,6,1,4}{%
      \ifnum\g<#3\relax
        \ifx#4d\def\hycell{hy/cell/draft}\else\def\hycell{hy/cell/rewritten}\fi
      \else\ifnum\g=#3\relax
        \ifx#4d\def\hycell{hy/cell/drafting}\else\def\hycell{hy/cell/rewriting}\fi
      \else
        \ifx#4d\def\hycell{hy/cell/mask}\else\def\hycell{hy/cell/draft}\fi
      \fi\fi
      \draw[\hycell] (\i*0.105,-0.15) rectangle ++(0.105,0.3);
    }
  \end{scope}}
\begin{adjustbox}{max width=\linewidth}
\begin{tikzpicture}[x=1cm,y=1cm,
  lab/.style={font=\sffamily\fontsize{7}{8.5}\selectfont,text=hymuted,align=center},
  ttl/.style={font=\sffamily\bfseries\fontsize{8}{9.5}\selectfont},
  lane/.style={font=\sffamily\bfseries\fontsize{9}{10.5}\selectfont,text=hyink,anchor=base west},
  tag/.style={rounded corners=2pt,font=\sffamily\bfseries\fontsize{7}{8.5}\selectfont,inner xsep=3pt,inner ysep=1.5pt,text=white}]
  \node[lane] at (0,0.42) {a\enspace Draft, then rewrite};
  \begin{scope}[on background layer]\node[hy/group,rounded corners=6pt,fit={(0,0)(10.6,-4.65)},inner sep=0pt] {};\end{scope}
  \node[ttl,anchor=west] at (0.25,-0.78) {Draft};
  \node[tag,fill=hywarm,anchor=west] at (0.25,-1.15) {\textit{T}\,=\,\TdraftVal};
  \node[lab,anchor=west] at (0.25,-1.5) {blind start};
  \hystrip{2.0}{-1.05}{0}{d}   \node[lab] at (2.84,-1.45) {round 0};
  \hystrip{3.95}{-1.05}{1}{d}  \node[lab] at (4.79,-1.45) {round 1};
  \node[lab,font=\sffamily\bfseries\fontsize{9}{10}\selectfont] at (6.0,-1.05) {$\cdots$};
  \hystrip{6.4}{-1.05}{7}{d}   \node[lab] at (7.24,-1.45) {round \the\numexpr\DraftRounds-1\relax};
  \draw[hy/arrow] (8.15,-1.05) -- (8.55,-1.05);
  \hystrip{8.65}{-1.05}{8}{d}  \node[lab] at (9.49,-1.45) {draft};
  \draw[hy/arrow,rounded corners=3pt] (9.49,-1.62) -- (9.49,-1.9) -- (1.8,-1.9) -- (1.8,-2.55) -- (1.94,-2.55);
  \node[ttl,anchor=west] at (0.25,-2.28) {Rewrite};
  \node[tag,fill=hyblue,anchor=west] at (0.25,-2.65) {\textit{T}\,=\,\TrewriteVal};
  \node[lab,anchor=west] at (0.25,-3.0) {group by group};
  \hystrip{2.0}{-2.55}{0}{w}   \node[lab] at (2.84,-2.95) {round 0};
  \hystrip{3.95}{-2.55}{1}{w}  \node[lab] at (4.79,-2.95) {round 1};
  \node[lab,font=\sffamily\bfseries\fontsize{9}{10}\selectfont] at (6.0,-2.55) {$\cdots$};
  \hystrip{6.4}{-2.55}{7}{w}   \node[lab] at (7.24,-2.95) {round \the\numexpr\RewriteRounds-1\relax};
  \draw[hy/arrow] (8.15,-2.55) -- (8.55,-2.55);
  \hystrip{8.65}{-2.55}{8}{w}  \node[lab] at (9.49,-2.95) {sequence};
  \foreach \x/\st/\t in {0.3/mask/masked, 1.75/drafting/drafting, 3.35/draft/draft token, 5.3/rewriting/rewriting, 7.15/rewritten/rewritten}{
    \draw[hy/cell/\st] (\x,-3.95) rectangle ++(0.14,0.28); \node[lab,anchor=west] at (\x+0.2,-3.81) {\t};}
  \node[lab,anchor=east] at (10.4,-4.32) {one encoder pass, \DecodePasses{} decoder passes};
  \node[lane] at (10.9,0.42) {b\enspace Equal diversity, higher scTM};
  \begin{scope}[on background layer]\node[hy/group,rounded corners=6pt,fit={(10.9,0)(16,-4.65)},inner sep=0pt] {};\end{scope}
  \begin{scope}[shift={(11.75,-3.55)}]
    \draw[line width=0.6pt,draw=hyink!60,-{Stealth[length=3pt,width=3pt]}] (0,0) -- (4.1,0);
    \draw[line width=0.6pt,draw=hyink!60,-{Stealth[length=3pt,width=3pt]}] (0,0) -- (0,3.1);
    \node[lab] at (2.05,-0.34) {sequence diversity};
    \node[lab,rotate=90] at (-0.3,1.6) {scTM};
    \draw[line width=1.5pt,hyblue,smooth] plot coordinates {(0.3,2.95) (1.1,2.85) (2.0,2.6) (2.9,2.1) (3.6,1.55)};
    \draw[line width=1.5pt,hyink!45,smooth] plot coordinates {(0.3,1.85) (1.1,1.7) (2.0,1.35) (2.9,0.8) (3.6,0.25)};
    \draw[dashed,line width=0.5pt,hyink!55] (1.5,0) -- (1.5,2.75);
    \fill[hyblue] (1.5,2.75) circle (0.075);
    \fill[hyink!55] (1.5,1.55) circle (0.075);
    \draw[<->,line width=0.7pt,hyblue] (1.5,1.68) -- (1.5,2.62);
    \node[lab,text=hyblue,font=\sffamily\bfseries\fontsize{7}{8.5}\selectfont,anchor=west,align=left] at (1.62,1.98) {higher\\[-1pt]scTM};
    \node[lab,anchor=west] at (1.6,0.3) {equal diversity};
    \node[lab,text=hyblue,font=\sffamily\bfseries\fontsize{7}{8.5}\selectfont,anchor=south west] at (0.2,3.05) {draft and rewrite};
    \node[lab,text=hyink!70,anchor=north west,align=left] at (0.22,1.42) {plain\\[-1pt]sampling};
  \end{scope}
\end{tikzpicture}
\end{adjustbox}
\caption{\textbf{Draft-and-rewrite sampling with decoupled temperatures.}
\textbf{a,} The positions are split into fixed groups (colors); a blind draft samples one group per round at the draft temperature $\Tdraft$, and the rewrite masks and resamples one group per round, with all other positions unmasked, at the rewrite temperature $\Trewrite$.
\textbf{b,} Schematic of the effect: at equal diversity, decoupled temperatures give a higher scTM than plain sampling; measured curves are in Figure~\ref{fig:teaser}c and Section~\ref{sec:ablation}.}
\label{fig:sampler}
\end{figure}
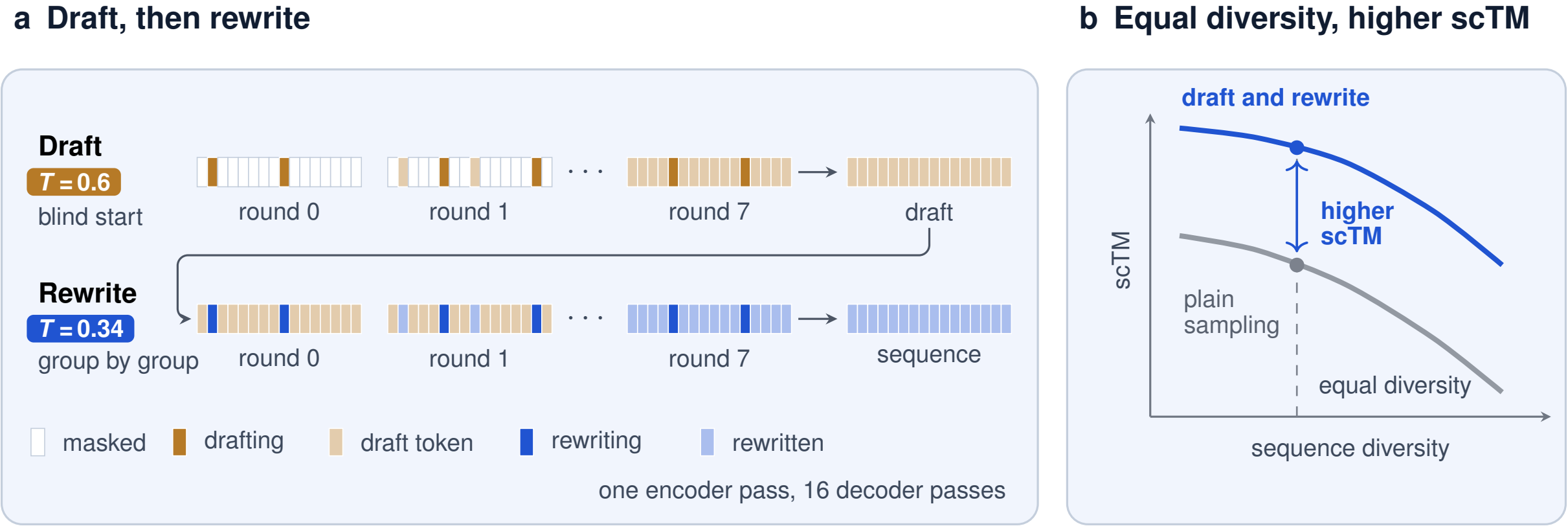

Because the decoder of Section~\ref{sec:arch} accepts a partially masked sequence, a sampler specifies which positions to sample and at what temperature.
The sampler encodes the backbone once and splits the positions at random into $\Rrounds=\DraftRounds$ fixed groups $G_0,\dots,G_{\Rrounds-1}$ of equal size (Figure~\ref{fig:sampler}a).
In the \emph{draft}, non-fixed positions start masked and round $r$ samples group $G_r$ conditioned on the observed sequence context.
In the \emph{rewrite}, round $r$ masks the non-fixed positions in $G_r$, keeps all other positions unmasked, and resamples that group.
Generation requires one forward pass through the encoder and \DecodePasses{} forward passes through the decoder (Algorithm~\ref{alg:rewrite} in Appendix~\ref{app:sampler}).

\subsection{Decoupled temperatures}
Sampling at a single temperature trades scTM for diversity.
\HyInv\ separates the two roles: the draft is sampled at a warm temperature $\Tdraft$, which sets the overall arrangement of the sequence, and the rewrite resamples each group at a cooler temperature $\Trewrite$, which makes every position consistent with its final context.
At equal diversity this gives a higher scTM than plain sampling at one temperature (Figure~\ref{fig:teaser}c): \GainDraftSixCathFour{} on CATH4.3 test and \GainDraftSixCplt{} on RecentPDB, measured on an earlier checkpoint of the same run (Appendix~\ref{app:sampler}).
The \HyInv\ default is the final checkpoint (update \RLStep) with $\Tdraft=\TdraftVal$ and $\Trewrite=\TrewriteVal$, fixed on the development panel and used for every result in this report.

\section{Efficient inference}\label{sec:efficiency}
\begin{figure}[tbp]
\centering\includegraphics{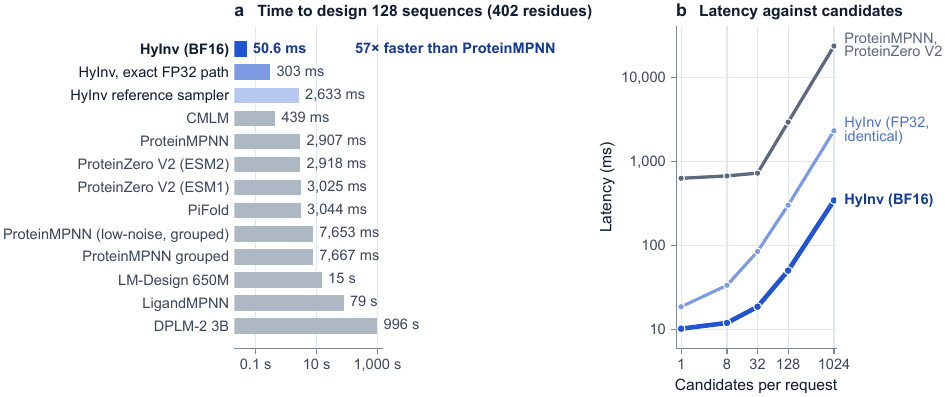}
\caption{\textbf{Inference time of \HyInv\ and the baselines.}
\textbf{a,} Time to design \EffNDesigns\ sequences for a \EffLenLong-residue backbone on one \EffGPU\ (logarithmic axis); baselines run with their released code at their published temperature.
\textbf{b,} Latency of one request against the number of candidates.}
\label{fig:eff_bars}
\end{figure}

\HyInv\ designs \EffNDesigns\ sequences for a \EffLenLong-residue backbone in \EffLatencyLong\ on one \EffGPU, \EffSpeedupLong$\times$ faster than the reference implementation of the sampler and \EffSpeedupMPNNLong$\times$ faster than ProteinMPNN (Table~\ref{tab:efficiency}, Figure~\ref{fig:eff_bars}).

Screening needs many candidates for one backbone, and sampling a design costs one encoder pass and \DecodePasses\ decoder passes (Section~\ref{sec:rewrite}), so the speed is set by how cheaply many candidates share these passes.
We reimplemented the sampler without changing the number of passes (Figure~\ref{fig:teaser}d).
Candidates of one request share a backbone, so the encoder state, the structure terms of the decoder and the first draft round are computed once for all of them.
The edge and node updates of every round run in a few fused kernels with \EffPrecision, so edge activations never reach memory, and each round is captured in a CUDA graph with its random numbers drawn ahead of time.
An exact FP32 variant returns the reference sequences.
A library of \EffLibN\ sequences for one backbone takes \EffLibTime\ (\EffThroughput).

\begin{table}[tbp]
\caption{\textbf{Inference time.}}
\label{tab:efficiency}
\centering\footnotesize
\setlength{\tabcolsep}{6pt}
\begin{tabular}{@{}lrrrr@{}}
\toprule
 & \multicolumn{2}{c}{128 designs (ms)} & 10,000 designs & Speed-up \\
\cmidrule(lr){2-3}
System & 148 res. & 402 res. & (s) & vs ProteinMPNN \\
\midrule
ProteinMPNN & 1{,}071 & 2{,}907 & 142.0 & {\color{hymuted}--} \\
ProteinZero V2 & 1{,}070 & 2{,}918 & 141.4 & 1.0$\times$ \\
HyInv, FP32 reference sampler & 999 & 2{,}633 & 129.1 & 1.1$\times$ \\
HyInv, exact FP32 path & 130 & 303 & 14.5 & 9.6$\times$ \\
\rowcolor[HTML]{DCE6FA}
\textbf{HyInv (BF16)} & \textbf{24.7} & \textbf{50.6} & \textbf{2.1} & \textbf{57$\times$} \\
\bottomrule
\end{tabular}
\tablenote{One NVIDIA H20, median of repeated runs, 128 designs per request; the 10,000-design time is for a 250-residue backbone. Timed with the supervised-stage weights (same architecture) and a rewrite temperature of \EffTimingTr. The exact FP32 path returns the same sequences as the reference sampler; the BF16 path (HyInv) changes the arithmetic slightly (Appendix~\ref{app:efficiency}).}

\end{table}

\paragraph{Quality.}
All sequences evaluated in Section~\ref{sec:results} come from the reference sampler, which the exact FP32 path reproduces sequence by sequence.
The BF16 path changes the arithmetic slightly: on the development panel its scTM at the default diversity differs from the reference sampler by \EffSadDelta, with diversity within \EffDivAgree, measured on an earlier RL checkpoint with the same architecture and kernels (Appendix~\ref{app:efficiency}).

\section{Results}\label{sec:results}
\subsection{Evaluation protocol}
\paragraph{Panels.}
Five test panels: CATH4.3 test (\NTargetsCathFour\ chains), CATH4.2 test (\NTargetsCathTwo)~\citep{ingraham2019}, TS500 (\NTargetsTSLarge) and TS50 (\NTargetsTSSmall)~\citep{gao2023}, and RecentPDB, \NTargetsCplt\ chains from experimental structures of recent PDB entries (\RecentPDBEntries{} entries, \RecentPDBResidues{} residues, lengths \RecentPDBLenMin--\RecentPDBLenMax), all standard chains with complete backbones, screened for homology to PDB and CATH entries before \RecentPDBCutoff.
ESMFold2~\citep{esmfold2} folds every design; RecentPDB is also folded by Boltz2~\citep{boltz2} and Protenix2~\citep{protenixv2}, which are not used in training.
\paragraph{Metrics.}
scTM, the TM-score between the predicted structure of a design and the input backbone, is the primary metric; native sequence recovery and sequence diversity (mean pairwise difference between the designs of a target) are reported beside it (Appendix~\ref{app:metrics}).
\paragraph{Baselines.}
Ten released models, each at its published single-temperature setting: ProteinMPNN and two grouped variants~\citep{dauparas2022}, ProteinZero V2 with ESM1 and ESM2 weights~\citep{wang2025proteinzero}, PiFold~\citep{gao2023}, CMLM~\citep{cmlm}, LM-Design 650M~\citep{zheng2023}, DPLM-2 3B~\citep{wang2024dplm2} and LigandMPNN~\citep{dauparas2023atomic}.
\paragraph{HyInv.}
One setting is used throughout: the final checkpoint (update \RLStep) with draft temperature \TdraftVal\ and rewrite temperature \TrewriteVal.
The checkpoint and temperatures were chosen on the development panel (CATH4.3 validation) and every test panel was evaluated once.
All evaluated sequences come from the reference sampler, which the exact FP32 path reproduces; the BF16 path was checked on the development panel (Section~\ref{sec:efficiency}).
Baselines are single published settings, so comparisons are at comparable rather than exactly equal diversity, and a repeated evaluation varies slightly (Appendix~\ref{app:protocols}).

\subsection{Comparison with published baselines}
\begin{figure}[tbp]
\centering\includegraphics{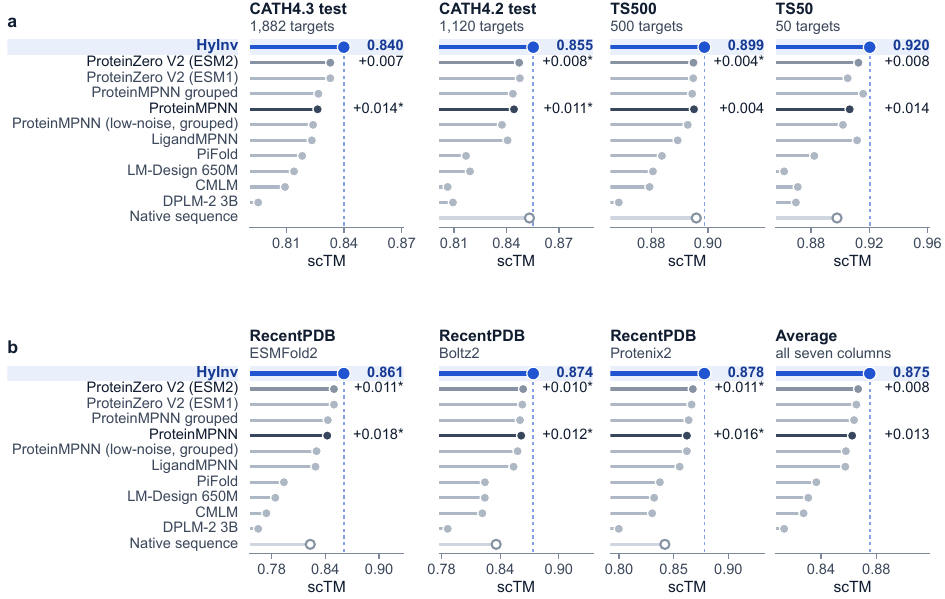}
\caption{\textbf{scTM of \HyInv\ and \BaseNBaselines\ published baselines.}
\textbf{a,} Four test panels with ESMFold2.
\textbf{b,} RecentPDB with ESMFold2, Boltz2 and Protenix2, and the average over the seven columns.
Methods are sorted by average scTM and the dashed line marks \HyInv.
At the right of each panel: scTM of \HyInv\ and its difference to ProteinMPNN and ProteinZero V2; a star marks a clearly positive difference (Appendix~\ref{app:protocols}).
Native: scTM of the native sequences.}
\label{fig:main_sctm}
\end{figure}

\begin{figure}[tbp]
\centering\includegraphics{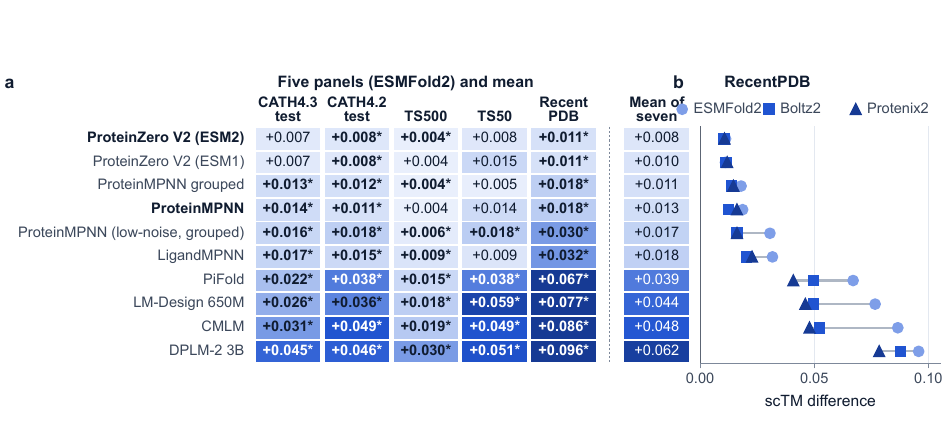}
\caption{\textbf{\HyInv\ minus baseline.}
\textbf{a,} Difference in scTM on the five panels (ESMFold2) and in the average over the seven columns; darker is larger, a star marks a clearly positive difference.
\textbf{b,} The same difference on RecentPDB under three structure predictors.}
\label{fig:delta}
\end{figure}

Figure~\ref{fig:main_sctm} compares scTM.
\HyInv\ has the highest scTM in all \MainNCols\ panel--predictor columns, and its average is \MainAvgScTMHyInv, against \MainAvgScTMPZ\ for ProteinZero V2 (ESM2) and \MainAvgScTMMPNN\ for ProteinMPNN.
The margin is smallest over ProteinZero V2 (ESM2) (\MainGainMin) and largest over DPLM-2 3B (\MainGainMax).
Per panel (Figure~\ref{fig:delta}, Table~\ref{tab:main}), \HyInv\ is above ProteinMPNN on every panel and comparable to or higher than ProteinZero V2; a star marks a clearly positive difference.
These gains come at comparable or higher diversity (\DivCathFour\ on CATH4.3 test against \BaseDivMPNNCathFour\ for ProteinMPNN and \BaseDivPZCathFour\ for ProteinZero V2; on RecentPDB \HyInv\ is \DivDeficitCpltPZ\ below ProteinZero V2) and with native recovery close to the baselines.
Parallel decoders (PiFold, CMLM, LM-Design, DPLM-2) sample at much lower diversity and have the lowest scTM.

\begin{table}[tbp]
\caption{\textbf{\HyInv\ and the two reference baselines on the five test panels.}}
\label{tab:main}
\centering\footnotesize
\setlength{\tabcolsep}{5pt}
\begin{tabular}{@{}lrrrrr@{}}
\toprule
Method & CATH4.3\,test & CATH4.2\,test & TS500 & TS50 & RecentPDB \\
\midrule
\multicolumn{6}{@{}l}{\textit{scTM}} \\
\addlinespace[1pt]
\rowcolor[HTML]{DCE6FA}
\textbf{HyInv} & \textbf{0.840} & \textbf{0.855} & \textbf{0.899} & \textbf{0.920} & \textbf{0.861} \\
ProteinZero V2 & 0.833 & 0.847 & 0.895 & 0.913 & 0.850 \\
ProteinMPNN & 0.826 & 0.844 & 0.895 & 0.907 & 0.842 \\
\midrule
\multicolumn{6}{@{}l}{\textit{scTM difference (HyInv minus method)}} \\
\addlinespace[1pt]
ProteinZero V2 & \textcolor[HTML]{2054D1}{$\blacktriangle$\,0.007}$^{\hphantom{\star}}$ & \textcolor[HTML]{2054D1}{$\blacktriangle$\,0.008}$^{\star}$ & \textcolor[HTML]{2054D1}{$\blacktriangle$\,0.004}$^{\star}$ & \textcolor[HTML]{2054D1}{$\blacktriangle$\,0.008}$^{\hphantom{\star}}$ & \textcolor[HTML]{2054D1}{$\blacktriangle$\,0.011}$^{\star}$ \\
ProteinMPNN & \textcolor[HTML]{2054D1}{$\blacktriangle$\,0.014}$^{\star}$ & \textcolor[HTML]{2054D1}{$\blacktriangle$\,0.011}$^{\star}$ & \textcolor[HTML]{2054D1}{$\blacktriangle$\,0.004}$^{\hphantom{\star}}$ & \textcolor[HTML]{2054D1}{$\blacktriangle$\,0.014}$^{\hphantom{\star}}$ & \textcolor[HTML]{2054D1}{$\blacktriangle$\,0.018}$^{\star}$ \\
\midrule
\multicolumn{6}{@{}l}{\textit{Sequence diversity}} \\
\addlinespace[1pt]
\rowcolor[HTML]{DCE6FA}
\textbf{HyInv} & 0.331 & 0.328 & 0.266 & 0.283 & 0.309 \\
ProteinZero V2 & 0.315 & 0.317 & 0.263 & 0.280 & 0.324 \\
ProteinMPNN & 0.314 & 0.312 & 0.264 & 0.272 & 0.307 \\
\midrule
\multicolumn{6}{@{}l}{\textit{Native sequence recovery}} \\
\addlinespace[1pt]
\rowcolor[HTML]{DCE6FA}
\textbf{HyInv} & 0.373 & 0.382 & 0.458 & 0.440 & 0.421 \\
ProteinZero V2 & 0.379 & 0.390 & 0.460 & 0.435 & 0.415 \\
ProteinMPNN & 0.387 & 0.398 & 0.466 & 0.445 & 0.428 \\
\bottomrule
\end{tabular}
\tablenote{ESMFold2; designs of a target are averaged first. $\Delta$ is HyInv minus the method: \textcolor[HTML]{2054D1}{$\blacktriangle$} HyInv higher, \textcolor[HTML]{AF4358}{$\blacktriangledown$} lower; $^{\star}$ clearly positive (Appendix~\ref{app:protocols}). Bold: best scTM of the column. ProteinZero V2 uses ESM2 weights; all baselines are in Appendix~\ref{app:results}.}

\end{table}

\subsection{Independent structure predictors}
ESMFold2 is also the reward predictor of RL, so RecentPDB is folded in addition by Boltz2 and Protenix2, which never enter training.
\HyInv\ has the highest scTM under both (Figure~\ref{fig:delta}b; one panel, protocol in Appendix~\ref{app:protocols}).

\subsection{Ablation study}\label{sec:ablation}
\begin{figure}[tbp]
\centering\includegraphics{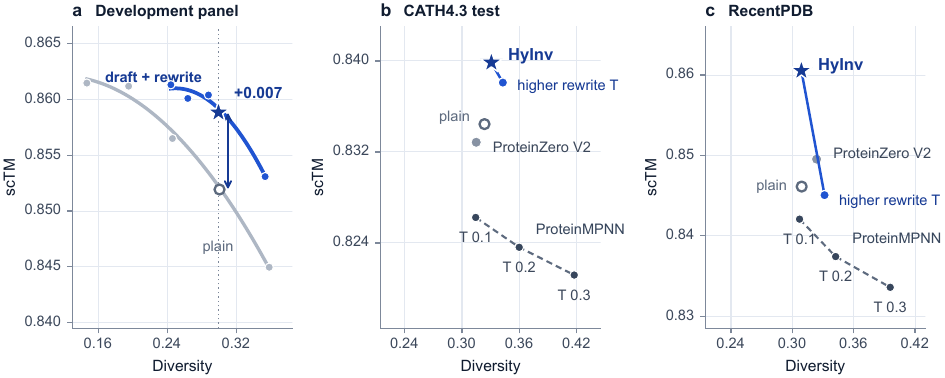}
\caption{\textbf{Draft-and-rewrite sampling at equal diversity.}
\textbf{a,} Development panel, final checkpoint: plain sampling (gray) and draft-and-rewrite (blue), four temperatures each; star: \HyInv, open circle: plain sampling at equal diversity.
\textbf{b, c,} CATH4.3 test and RecentPDB against ProteinMPNN at three temperatures (dashed) and ProteinZero V2; the extra blue point is a higher rewrite temperature.}
\label{fig:ablation}
\end{figure}

\begin{figure}[tbp]
\centering\includegraphics{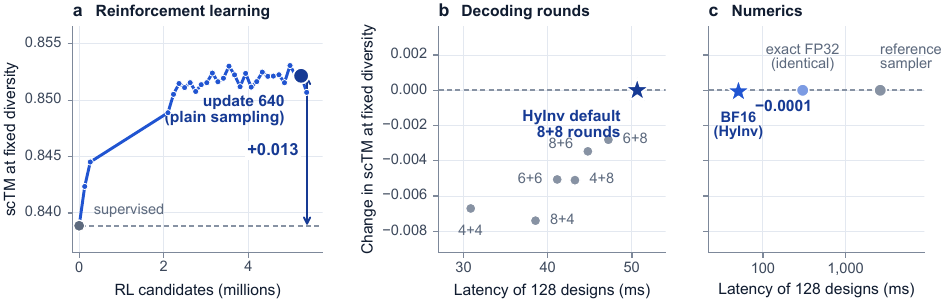}
\caption{\textbf{Reinforcement learning, decoding rounds and numerics.}
\textbf{a,} Development scTM at fixed diversity ($D=\DevD$) against the number of RL candidates, plain sampling; gray: supervised checkpoint, large dot: update \RLStep.
\textbf{b,} Change in scTM at fixed diversity and latency when the number of draft and rewrite rounds is reduced from the default.
\textbf{c,} Exact FP32 and BF16 path: latency and change in scTM at fixed diversity.}
\label{fig:ablation_rl}
\end{figure}

We vary one component of \HyInv\ at a time.
\emph{Sampler.} Draft-and-rewrite sampling gives a higher scTM than plain sampling at equal diversity on the development panel and on both held-out panels (Figure~\ref{fig:ablation}).
\emph{Temperature.} A higher rewrite temperature moves \HyInv\ along its curve toward more diversity and slightly lower scTM; the default sits at development diversity \DevD.
\emph{Reinforcement learning.} RL raises scTM at fixed diversity from \DevSadSFT\ to \DevSadRL\ with plain sampling (\DevGainRL, Figure~\ref{fig:ablation_rl}a).
\emph{Decoding rounds.} All \DecodePasses\ passes give the best scTM at fixed diversity; fewer rounds are faster at lower scTM (Figure~\ref{fig:ablation_rl}b).
\emph{Numerics.} The exact FP32 path and the BF16 path differ in scTM at fixed diversity by \EffSadDelta\ on the development panel (Figure~\ref{fig:ablation_rl}c).

\section{Applications}\label{sec:applications}
The same \HyInv\ default setting serves four single-chain design tasks without task-specific tuning.
All structures shown are ESMFold2 predictions of \HyInv\ designs; protocols and selection rules are in Appendix~\ref{app:applications}.

\begin{figure}[tbp]
\centering\includegraphics{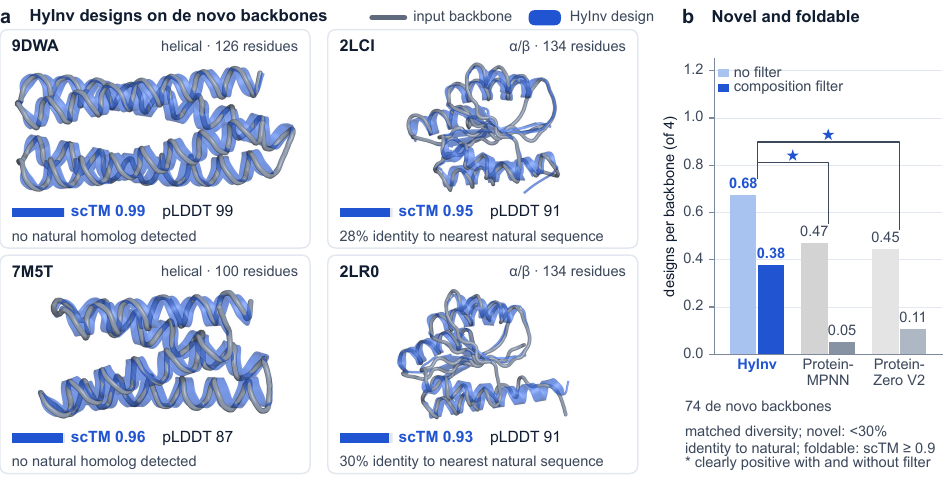}
\caption{\textbf{De novo backbones.}
\textbf{a,} \HyInv\ designs (blue ribbons) on the input backbones (gray); the best design of each of four backbones.
\textbf{b,} Novel and foldable designs per backbone at matched diversity, without (light) and with (dark) the composition filter (entropy $\ge\DemoFiltEntropy$ bits, hydrophobic fraction \DemoFiltHydLo--\DemoFiltHydHi, homopolymer run $\le\DemoFiltHomopolymer$).}
\label{fig:demo_denovo}
\end{figure}

\paragraph{De novo backbones.}
On \DemoDenoN\ de novo backbones, \HyInv\ yields more designs that are both novel (below \DemoDenoNovelThresh\ identity to known natural sequences) and foldable (scTM of at least \DemoDenoFoldThresh) than ProteinMPNN or ProteinZero V2 at matched diversity, with and without a filter on sequence composition (entropy of at least \DemoFiltEntropy{} bits, hydrophobic fraction \DemoFiltHydLo--\DemoFiltHydHi, longest homopolymer run at most \DemoFiltHomopolymer), since low-complexity sequences can be folded to a high scTM (Figure~\ref{fig:demo_denovo}): \DemoDenoYieldU\ unfiltered and \DemoDenoYield\ filtered per backbone, against \DemoDenoYieldMPNNU\ and \DemoDenoYieldMPNN\ for ProteinMPNN and \DemoDenoYieldPZU\ and \DemoDenoYieldPZ\ for ProteinZero V2.

\begin{figure}[tbp]
\centering\includegraphics{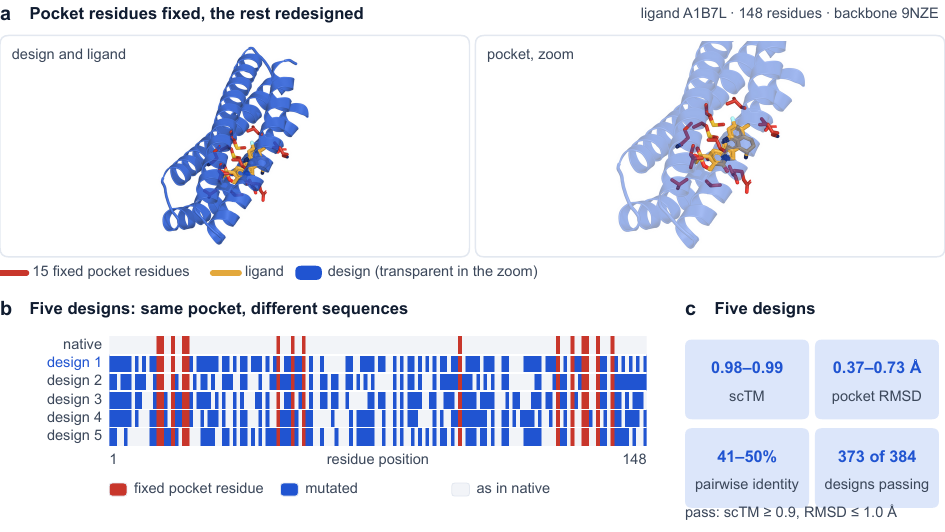}
\caption{\textbf{Pocket-constrained sequence library.}
\textbf{a,} A design with the pocket residues fixed to their native identity (red) next to the ligand of the native structure.
\textbf{b,} Five passing designs picked by maximum dissimilarity: the pocket is identical, the rest differs.}
\label{fig:demo_pocket}
\end{figure}

\paragraph{Pocket-constrained libraries.}
With the \DemoPocketNFixed\ ligand-contacting residues fixed, \HyInv\ redesigns the rest of a \DemoPocketLen-residue binder into diverse sequences that keep the pocket: \DemoPocketPass\ of \DemoPocketN\ designs pass on this backbone, which is shown for display (the mean pass rate over \DemoPocketNBack\ pocket backbones is \DemoPocketMeanPass; Figure~\ref{fig:demo_pocket}).

\begin{figure}[tbp]
\centering\includegraphics{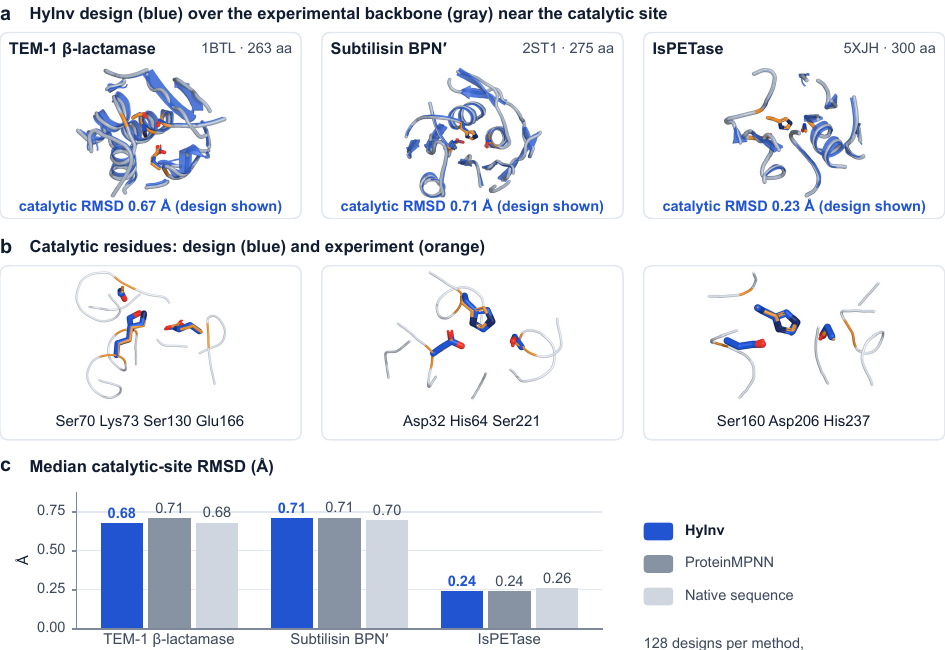}
\caption{\textbf{Enzymes with fixed catalytic residues.}
\textbf{a,} Designs (blue) over the experimental structures (gray), catalytic residues in red.
\textbf{b,} Catalytic site of the design (blue) and of the experiment (orange).
\textbf{c,} Median heavy-atom RMSD of the catalytic residues for \DemoEnzN\ designs of \HyInv\ and of ProteinMPNN and for the natural sequence. In a and b, the design with the highest scTM of \DemoEnzN.
The enzymes are known proteins: the figure shows that the catalytic geometry is kept.}
\label{fig:demo_enzyme}
\end{figure}

\paragraph{Enzymes with the catalytic site kept.}
With the catalytic residues fixed, \HyInv\ redesigns TEM-1 $\beta$-lactamase, subtilisin and PETase and keeps the predicted catalytic geometry at the level of the natural sequence (Figure~\ref{fig:demo_enzyme}).

\begin{figure}[tbp]
\centering\includegraphics{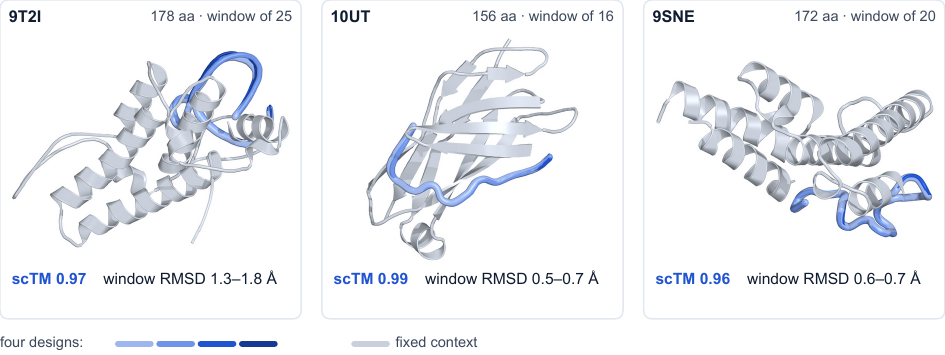}
\caption{\textbf{Loop redesign.}
The four highest-scTM designs of \DemoLoopPool\ (colored tubes) for a window with the rest of the structure fixed (gray); RMSD is the C$\alpha$ RMSD of the window to the experimental structure.}
\label{fig:demo_loop}
\end{figure}

\paragraph{Loop redesign.}
\HyInv\ redesigns loop windows of \DemoLoopWinLo\ to \DemoLoopWinHi\ residues in a fixed context, with the window within \DemoLoopRMSDLo\ to \DemoLoopRMSDHi\ of the experimental structure (Figure~\ref{fig:demo_loop}).

\section{Conclusion}
\HyInv\ is a \ParamsAdj{} graph network trained with random-order supervision and structure-reward reinforcement learning, deployed with a draft-and-rewrite sampler whose draft and rewrite temperatures are set separately.
On five benchmark panels, one of them under three structure predictors, it reaches a higher average scTM than all \BaseNBaselines\ external baselines at comparable or higher sequence diversity, and an optimized implementation designs \EffNDesigns\ sequences for a \EffLenLong-residue backbone in \EffLatencyLong.
The same default setting supports de novo backbones, pocket-constrained libraries, enzyme redesign and loop redesign.
\HyInv\ targets single-chain backbones; designing for multi-chain interfaces and membrane environments is a natural next step.

\FloatBarrier
\section*{Author Information}
\phantomsection\addcontentsline{toc}{section}{Author Information}
Haowei Lin, Cai Zhou and Xiangxin Zhou. All authors are affiliated with Tencent Hunyuan.

{\small\setlength{\bibsep}{2pt}
\bibliographystyle{plainnat}
\bibliography{refs}}

\clearpage
\appendix

\newcommand{\AppxRefoldMeanDiff}{0.0003}
\newcommand{\AppxRefoldMaxDiff}{0.0024}
\newcommand{\AppxRefoldNCells}{69}
\newcommand{\AppxDenovoN}{74}
\newcommand{\AppxDenovoClusters}{58}
\newcommand{\AppxNativeControlN}{193}
\newcommand{\AppxRegionNStrict}{8}
\newcommand{\AppxRegionNNamed}{4}
\newcommand{\AppxRegionDesigns}{24{,}672}
\newcommand{\AppxRegionPerCell}{128}
\newcommand{\AppxLibN}{12}
\newcommand{\AppxLibPerCell}{384}
\newcommand{\AppxLibPlates}{200}
\newcommand{\AppxLibPlateSmall}{96}
\newcommand{\AppxLibPlateLarge}{384}
\newcommand{\AppxLibScTMThresh}{0.90}
\newcommand{\AppxLibSiteThresh}{1.0}
\small
\setlength{\parskip}{0pt}
\section{Evaluation protocol and statistics}\label{app:protocols}
\small
This appendix defines the differences and stars of Section~\ref{sec:results} and records how the checkpoint and the sampler setting were fixed.

\paragraph{Panels.}
Table~\ref{tab:protocol} lists the panels.
CATH4.3 validation is the development panel: it fixed the sampler setting and the checkpoint and monitored RL training; each test panel was evaluated once for the final comparison.
RecentPDB consists of \NTargetsCplt{} experimental chains from \RecentPDBEntries{} recent PDB entries (\RecentPDBResidues{} residues, lengths \RecentPDBLenMin--\RecentPDBLenMax), all standard chains with complete backbones and no length truncation, screened for homology to PDB and CATH entries before \RecentPDBCutoff{} and fixed before evaluation, not selected by model performance.
A residue is a \emph{valid site} if its N, C$\alpha$, C and O atoms are present and its native residue is a standard amino acid; all scores are computed over valid sites.
The resampling unit of the paired intervals is the CATH topology cluster on the two CATH panels and the single chain elsewhere.

\begin{table}[!htbp]
\caption{\textbf{Evaluation panels.}}
\label{tab:protocol}
\centering\small
\begin{tabular}{@{}lrrll@{}}
\toprule
Panel & Chains & Clusters & Structure predictors & Use \\
\midrule
CATH4.3 test & \NTargetsCathFour & \NClustersCathFour & ESMFold2 & comparison; sampler check \\
CATH4.2 test & \NTargetsCathTwo & \NClustersCathTwo & ESMFold2 & comparison \\
TS500 & \NTargetsTSLarge & \NClustersTSLarge & ESMFold2 & comparison \\
TS50 & \NTargetsTSSmall & \NClustersTSSmall & ESMFold2 & comparison \\
RecentPDB & \NTargetsCplt & \NClustersCplt & ESMFold2, Boltz2, Protenix2 & comparison; sampler check \\
\midrule
CATH4.3 validation & \NTargetsDev & \NClustersDev & ESMFold2 & development \\
\bottomrule
\end{tabular}
\end{table}

\paragraph{Folding and seeds.}
\begin{itemize}\setlength\itemsep{1pt}
\item ESMFold2~\citep{esmfold2} runs with its fused backend, \FoldLoops{} recycling loops, \FoldDiffusionSteps{} diffusion steps and one structure per sequence, in FP32 with TF32 off; the input is the designed sequence with sequence-derived templates only, and native coordinates enter only through TM-align. The accelerated folder of Appendix~\ref{app:training} serves only the RL reward.
\item scTM, recovery and diversity come from the scoring run; pLDDT, scRMSD and lDDT (Appendix~\ref{app:metrics}) come from a second fold of the same sequences with the confidence head on, whose scTM agrees with the scoring run to \AppxRefoldMeanDiff{} on average over \AppxRefoldNCells{} cells. The reported scTM is always the scoring value.
\item Fold seeds derive from the chain identifier on the development panel; on the public panels with $K=\KRef$ designs per chain they are the seeds recorded with the baseline samples, so comparisons with baseline sequences share the folding randomness. For $K=\KHeld$ they are hashes of a base seed, the chain and the batch chunk.
\item Boltz2~\citep{boltz2} and Protenix2~\citep{protenixv2} run on RecentPDB only, with the alignment built from the designed sequence by MMseqs2~\citep{steinegger2017mmseqs2} against UniRef50; these scores partly reflect resemblance to a natural family and serve as a robustness check.
\end{itemize}

\paragraph{Scores.}
For a chain with valid-site set $\mathcal V$ and designs $a^{1},\dots,a^{K}$, scTM is the mean TM-align score~\citep{zhang2005} of the predicted against the native C$\alpha$ trace at $\mathcal V$, normalized by $|\mathcal V|$ and clipped at one, and the diversity is
\begin{equation}
\Div \;=\; \frac{2}{K(K-1)}\sum_{j<k}\frac{1}{|\mathcal V|}\sum_{i\in\mathcal V}\mathbb 1\!\left[a^{j}_i\neq a^{k}_i\right].
\end{equation}
Panel values are macro-averages over chains.
Diversity is counted at valid sites for every method: baseline sequences carry X at invalid sites, and counting them as identical would lower their diversity (ProteinMPNN on CATH4.3 test: \BaseDivAllMPNNCathFour{} at all sites, \BaseDivMPNNCathFour{} at valid sites).

\paragraph{Paired differences and intervals.}
A paired difference is the macro-average of the per-chain scTM difference between HyInv and a baseline.
Its 95\% interval is the percentile interval of \BootB{} resamples of the resampling unit with one resample shared by all rows compared and a fixed seed (\MetSeed).
A difference carries a star in the figures and tables when its interval excludes zero; a few starred intervals have a lower end within $0.001$ of zero and are borderline (Tables~\ref{tab:full_sctm} and~\ref{tab:delta_full}).
Intervals are not adjusted for multiple comparisons.
Recovery, diversity, pLDDT, scRMSD and lDDT differences use the same resamples.
Tables~\ref{tab:full_sctm} and~\ref{tab:delta_full} give all scores and differences with their intervals.

\paragraph{Scores at equal diversity.}
A sampler family is evaluated at a grid of rewrite temperatures, giving points of diversity and scTM on the same chains.
The scTM at fixed diversity $D$ is the value at $D$ of the least-squares quadratic fit $s=\alpha+\beta d+\gamma d^{2}$,
\begin{equation}
\SaD \;=\; \alpha+\beta D+\gamma D^{2},
\end{equation}
with no extrapolation.
The gain of a setting is its $\SaD$ minus that of the unmodified sampler of the same checkpoint on the same panel, with an interval from \SadB{} draws in which all points of all curves are recomputed from one cluster resample.
The grids are $\{0.3,0.4,0.5,0.7\}$ for the unmodified sampler, $\{0.1,0.2,0.3,0.5\}$ for the decoupled sampler at $\Tdraft=\TdraftVal$ and $\{0.3,0.4,0.5,0.7\}$ at $\Tdraft=\TdraftAltVal$; on the held-out panels $D$ is the development value times the ratio of the panel to the development diversity of the unmodified sampler.
HyInv has diversity above ProteinMPNN with a clear interval on \MetDivAboveMPNNList{} and above ProteinZero V2 (ESM2) on \MetDivAbovePZList, and \DivDeficitCpltPZ{} below ProteinZero V2 (ESM2) on RecentPDB; at the slope of the development curve (\DevSlope{} scTM per unit of diversity) the largest offset, \DivMaxExcess, is worth at most \DivCostMax{} scTM, so the comparisons are close to, but not exactly at, equal diversity.

\paragraph{Checkpoint and sampler setting.}
Both were chosen on the development panel (CATH4.3 validation) alone.
The checkpoint is the latest RL snapshot whose development S@\SadRefDev{} is within \RLTieWindow{} of the best (update \RLStep); the best-scoring snapshot differs from it by \SelGap, below the noise of one evaluation.
The default sampler setting (draft \TdraftVal, rewrite \TrewriteVal, development diversity \DevD) was read off the development curve before the test panels were generated with this checkpoint.
The decoupled draft was then confirmed on held-out panels with new draws (Appendix~\ref{app:sampler}); because the draft temperature was chosen on the development panel, the held-out confirmation is the evidence for the sampler gain.
Each test panel was evaluated once.
The development panel also served to choose the RL learning rate and training configuration.

\paragraph{Run-to-run variation.}
Two independent generate-and-fold runs of the same policy and sampler on the development panel differ by \NoisePairedDiffs{}, so one evaluation carries noise of about \NoiseSingleEval{}; refolding identical sequences with other fold seeds moves a panel mean by less than \NoiseFoldOnlyMax{}.
Differences between checkpoints below about $0.003$ are not resolvable.

\section{Metric definitions}\label{app:metrics}
All quantities are measured on predicted structures and designed sequences; none is an experimental measurement.
Panel-level values average the designs of a chain first and then the chains (Appendix~\ref{app:protocols}).
\begin{description}[leftmargin=0pt,labelindent=0pt,itemsep=3pt,font=\bfseries]
\item[scTM.] TM-score~\citep{zhang2004scoring} of the predicted structure of a design to the native backbone, computed on the C$\alpha$ trace at valid sites, normalized by the native length and clipped at one; higher is better. A failed fold is a gap in the average, not imputed. This is the primary quantity and the RL reward.
\item[Recovery.] Fraction of valid sites at which the design carries the native residue. It is descriptive, not an objective: it has weight zero in the reward and selects no checkpoint, and methods with high recovery tend to have low diversity and lower scTM.
\item[Diversity.] Mean pairwise normalized Hamming distance between the $K$ designs of a chain over valid sites (\Div{} of Appendix~\ref{app:protocols}). It is a property of the design set and is matched rather than ranked.
\item[pLDDT.] The folder's mean per-residue confidence in its own prediction over valid sites, on a scale from 0 to 100. It comes from the same folder as scTM and is not compared across predictors.
\item[scRMSD.] Self-consistency RMSD in {\AA}: RMSD between the C$\alpha$ atoms of the predicted structure of a design and the native backbone at valid sites, after optimal (Kabsch) superposition with the residue correspondence fixed by position; lower is better.
\item[lDDT.] C$\alpha$-only lDDT-style score without superposition: over all C$\alpha$ pairs at distance below $15$~{\AA} in the native structure, the fraction whose distance error is below $0.5$, $1$, $2$ and $4$~{\AA}, averaged over the four thresholds, with the pairs pooled over the whole design; between 0 and 1, higher is better.
\item[Site and region RMSD.] For the application tasks (Appendix~\ref{app:applications}): the C$\alpha$ RMSD at the fixed pocket or catalytic residues after one global fit on all valid C$\alpha$ atoms (site RMSD), and the C$\alpha$ RMSD of a redesigned loop or window after fitting the unchanged context (region RMSD); the catalytic-site RMSD uses all heavy atoms of the catalytic residues.
\item[Novel-and-foldable yield.] Number of designs per backbone with scTM above the foldability threshold and a sequence identity to the nearest UniRef50 and pretraining-library sequence below the novelty threshold (both stated in Appendix~\ref{app:applications}).
\item[Distinct passing designs.] Number of different sequences in a plate of designs that pass the structure criteria of the pocket task, counted after clustering at 50\% sequence identity.
\end{description}

\section{Model, data and training details}\label{app:training}

\paragraph{Configuration.}
The network has \ParamsExact{} parameters: width \Width, \EncLayers{} encoder and \DecLayers{} decoder layers, \Neighbours{} nearest neighbors, random initialization.
Table~\ref{tab:config} lists the two training stages.

\begin{table}[!htbp]
\caption{\textbf{Training configuration.}}
\label{tab:config}
\centering\small
\begin{tabular}{@{}lll@{}}
\toprule
 & Supervised & RL \\
\midrule
Data & \SftSeqs{} structures & \PoolSize{} backbones \\
Objective & random-order likelihood & reward, leave-one-out advantage \\
Updates & \SftUpdates{} & \RLStep{} (full run \RunUpdates{}) \\
Batch & window up to \SftMaxLen{} & \CohortSize{} backbones $\times$ \GroupSize{} \\
Optimizer & AdamW, $\beta=$ \SftBetas & AdamW, $\beta=$ \RLBetas \\
Learning rate & Noam schedule & \RLLR{} constant \\
Regularization & label smoothing \SftLabelSmoothing & KL \KLInit{}, then \KLLate \\
Sampling & -- & temperature \RLTemp \\
\bottomrule
\end{tabular}
\end{table}

\paragraph{Reward and stability proxy.}
scTM is computed with ESMFold2 and TM-align as in Appendix~\ref{app:protocols}, normalized by the native valid length and clipped at one.
The stability proxy $s$ is the bounded likelihood contrast of ProteinMPNN~\citep{dauparas2022} with the weights of the ProteinMPNN baseline: with $\ell_X(a)$ the summed negative log-likelihood of sequence $a$ at the $N$ valid sites given the backbone, $\ell_0(a)$ the same with zeroed coordinates (all four terms sharing one tied random decoding order) and $a^{\star}$ the native sequence,
\begin{equation}
\Delta(a)=0.593\Big[\big(\ell_X(a)-\ell_0(a)\big)-\big(\ell_X(a^{\star})-\ell_0(a^{\star})\big)\Big],\qquad
s(a)=\tfrac12-\tfrac12\tanh\!\frac{\Delta(a)}{0.593\,N\,0.5}\in(0,1).
\end{equation}
The native sequence enters only as the reference of the contrast; recovery has weight zero.
The term supplies five percent of the reward and is a model-derived proxy, not a stability measurement.

\paragraph{RL objective.}
The advantage of a candidate $i$ of a backbone uses all $n=\GroupSize$ candidates of that backbone~\citep{rloo},
\begin{equation}
A_i=\frac{1}{\AdvDivisor}\Bigl(r_i-\frac{1}{n-1}\sum_{j\neq i}r_j\Bigr),
\label{eq:rloo}
\end{equation}
which removes most of the dependence on how hard a backbone is, without a learned value network.
With $\ell_i=\frac{1}{L_i}\sum_{t}\log\pitheta^{\Troll}(a_{i,t}\mid c_{i,t})$ the length-normalized log-likelihood of candidate $i$ along its sampling order, $c_{i,t}=(\bb,\aa_{i,<t})$ the sampled context and $L_i$ the number of valid positions, the loss of one backbone, with the advantage of Eq.~\eqref{eq:rloo}, is
\begin{equation}
\mathcal L(\theta)=\frac1n\sum_{i=1}^{n}\Bigl[-A_i\,\ell_i+\beta\,\frac{1}{L_i}\sum_{t}\mathrm{KL}\bigl(\pitheta^{\Troll}(\cdot\mid c_{i,t})\,\big\Vert\,\piref^{\Troll}(\cdot\mid c_{i,t})\bigr)\Bigr],
\end{equation}
where $\piref$ is the fixed supervised checkpoint and $\beta$ is the KL coefficient of Table~\ref{tab:config}.

\begin{figure}[t]
\centering\includegraphics[width=\linewidth]{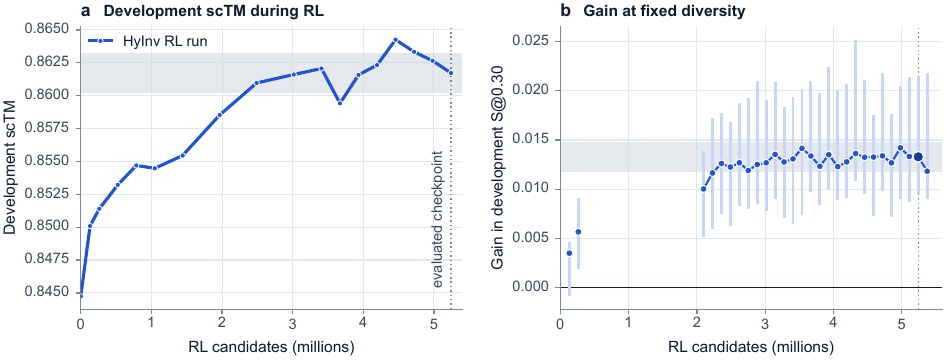}
\caption{\textbf{RL training on the development panel.}
\textbf{a,} Monitoring scTM during training (confidence sampler, fixed protocol).
\textbf{b,} Gain in scTM at equal diversity over the supervised checkpoint; the large marker is the evaluated checkpoint (update \RLStep), bars are 95\% intervals over clusters.}
\label{fig:rlcurve}
\end{figure}

\paragraph{Optimization and parallel execution.}
\begin{itemize}\setlength\itemsep{1pt}
\item Gradients are summed per backbone, reduced in double precision and divided by the cohort size, so the assignment of backbones to GPUs does not change the update; log-probabilities are recomputed in one causal pass given the stored order (largest rollout-to-replay difference: \ReplayErrMax).
\item Each backbone of an update goes to one whole GPU, assigned longest-processing-time-first on a cost that grows like length to the power $1.9$ (planned imbalance \PlanImbalance, observed \ObsImbalance).
\item The run was restarted from its last checkpoint with a different number of GPUs, keeping its update numbering: \GPUsEarly{} GPUs for updates 1 to \UpdLastEarly, \GPUsMid{} up to \UpdLastMid, \GPUsLate{} up to \UpdLastLate{} and \GPUsFinal{} from update \UpdFirstFinal{} on (median \StepSecEarly, \StepSecMid, \StepSecLate{} and \StepSecFinal{} seconds per update, each update with \CandPerUpdate{} candidates); the evaluated checkpoint (update \RLStep) was trained in the phase with \CkptGPUs{} GPUs.
\item The reward uses an accelerated ESMFold2 (confidence and output heads skipped, diffusion pair-bias terms and alignment features cached; each fold faster by \FoldSpeedup), which is not bit-identical to the evaluation model: on six development chains the largest coordinate difference is \EquivCoordDiff{} against \EquivSelfDiff{} between two runs of the original, and a seed-matched test on \EquivNSeqs{} sequences gives a median absolute scTM difference of \EquivMedianDScTM. All evaluation numbers use the original recipe.
\item Every few updates the trainer scores the development panel under a fixed monitoring protocol (confidence sampler at $T=0.1$, $K=\KRef$; Figure~\ref{fig:rlcurve}a), which is not comparable with the equal-diversity read-outs of Figure~\ref{fig:rlcurve}b.
\end{itemize}

\paragraph{Data and pool.}
\begin{itemize}\setlength\itemsep{1pt}
\item The supervised corpus has \SftSeqs{} unique sequences. Records need a valid-residue ratio of at least 0.95 and canonical residues only, predicted structures a mean pLDDT of at least 85. Records are grouped into components (30\%-identity protein-entity clusters joined by exact chain-sequence links across sources), the base weight of a record is the reciprocal of its component's size within its source type, and source weights give experimental structures about 70\% and predicted structures about 30\% of the expected loss mass.
\item The RL pool has \PoolSize{} backbones (\PoolExp{} experimental, \PoolPred{} predicted; mean length \PoolMeanLen, range 48 to \SftMaxLen) under the same quality rules, with at most one predicted record per cascaded 50\%-identity MMseqs2 cluster and at most two experimental records per cluster; candidates were drawn with weights that favor harder backbones.
\item The pool was filtered against the benchmark panels with a same-source homology mask built from MMseqs2 searches~\citep{steinegger2017mmseqs2} of the candidate records against the benchmark sequences: a record is excluded if it is positive under the mask, and records are used only if that search queried them. The supervised corpus was filtered by exact-sequence removal and a homology filter.
\item We do not claim zero sequence overlap between the training data and the evaluation panels.
\end{itemize}

\section{Baselines and full results}\label{app:results}

\subsection{Baseline settings}
The ten external baselines are published settings: ProteinMPNN, ProteinMPNN with grouped decoding, the same with a checkpoint trained with less coordinate noise (``low-noise, grouped''), ProteinZero V2 with the ESM1 and the ESM2 variant, PiFold, CMLM, LM-Design 650M, DPLM-2 3B and LigandMPNN.
Each has data on all \BaseNPanelsESM{} ESMFold2 panels and on RecentPDB also under Boltz2 and Protenix2, scored with the protocol of Appendix~\ref{app:protocols} at the single setting at which it was published, not tuned to the panels.
Table~\ref{tab:baselines} lists decoding and temperature.
ProteinZero V2 is declared for backbones of up to 100 residues and is run here on the full panels.
PiFold, CMLM and LM-Design decode all positions in parallel from one categorical distribution, and DPLM-2 3B decodes by argmax with deterministic unmasking, so its two designs of a chain are almost always the same sequence (diversity \BaseDivDPLMCathFour{} on CATH4.3 test); LigandMPNN is run on backbone coordinates only.

\begin{table}[!htbp]
\caption{\textbf{Baseline settings, weights and reproduction check.}}
\label{tab:baselines}
\centering
\footnotesize
\begin{tabular}{@{}llcllc@{}}
\toprule
Baseline & Decoding & $T$ & Regenerated on & $|\Delta$scTM$|$ & Same seq. \\
\midrule
ProteinMPNN & autoregressive & 0.1 & none & -- & -- \\
ProteinMPNN grouped & autoregressive, grouped & 0.1 & CATH4.3 test & 0.0009 & 100.0\% \\
ProteinMPNN (low-noise, grouped) & autoregressive, grouped & 0.1 & four panels & 0.0007 & 100.0\% \\
ProteinZero V2 (ESM1) & autoregressive & 0.1 & CATH4.3 test & 0.0006 & 100.0\% \\
ProteinZero V2 (ESM2) & autoregressive & 0.1 & none & -- & -- \\
PiFold & one parallel pass & 0.1 & four panels & 0.0004 & 100.0\% \\
CMLM & one parallel pass & 0.1 & four panels & 0.0022 & 100.0\% \\
LM-Design 650M & iterative denoising (5 steps) & 0.1 & four panels & 0.0005 & 100.0\% \\
DPLM-2 3B & argmax decoding (100 steps) & 0.1 & four panels & 0.0001 & 99.2\% \\
LigandMPNN & autoregressive & 0.1 & four panels & 0.0005 & 100.0\% \\
\bottomrule
\end{tabular}
\tablenote{All baselines use the released weights. The reproduction check regenerates the baseline on RecentPDB, folds it with ESMFold2 and compares the target-averaged scTM with the released value ($|\Delta$scTM$|$) and the fraction of identical sequences. DPLM-2 3B runs in BF16 autocast.}

\end{table}

Some ESMFold2 cells were regenerated for this report with the same weights, temperature and sampler; a regenerated cell is used only where the same procedure reproduces the released score on RecentPDB (difference \BaseGateDevMin{} to \BaseGateDevMax, within \BaseGate), and for DPLM-2 3B \BaseDPLMSeqIdent{} of the sequences are identical to the released ones.
The RecentPDB cells under all three predictors are released.
Folding seeds follow the released per-chain seeds, so cells are paired with HyInv at the chain level.
The ProteinMPNN temperatures other than the published one were generated for this report.

\subsection{All baselines on every panel}
\begin{figure}[t]
\centering\includegraphics[width=\linewidth]{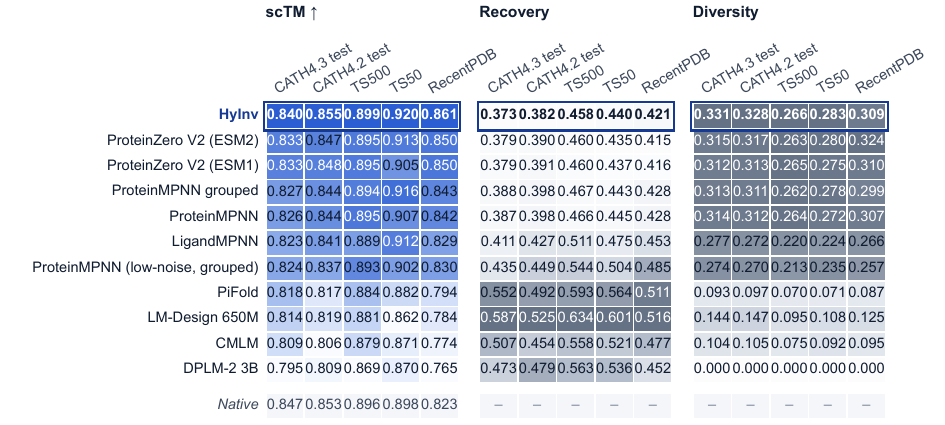}
\caption{\textbf{scTM, recovery and diversity of every method on the five ESMFold2 panels.}
Rows are sorted by mean scTM, HyInv first; columns are shaded within each panel (darker is better for scTM; recovery and diversity are shown in gray as properties).}
\label{fig:metrics_heat}
\end{figure}

\begin{figure}[t]
\centering\includegraphics[width=\linewidth]{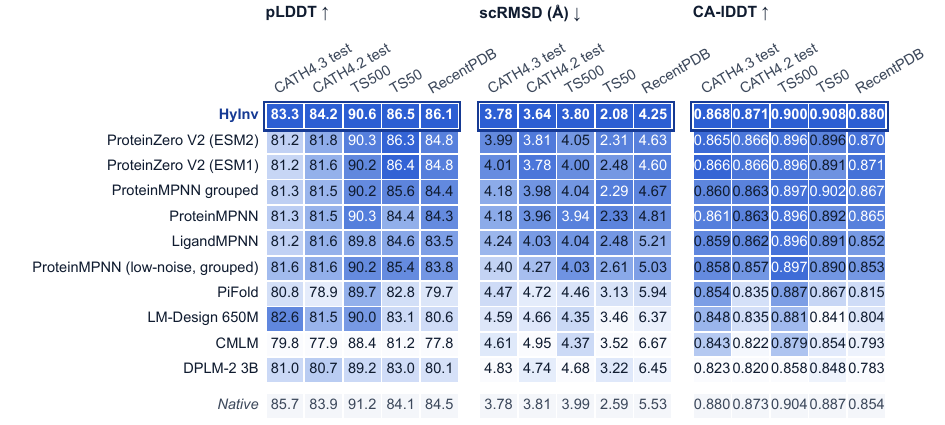}
\caption{\textbf{pLDDT, scRMSD and lDDT of every method on the five ESMFold2 panels.}
Same layout as Figure~\ref{fig:metrics_heat}; darker is better (higher pLDDT and lDDT, lower scRMSD).}
\label{fig:metrics_heat_structure}
\end{figure}

Table~\ref{tab:full_sctm} lists the scTM of every method on every panel and predictor, Table~\ref{tab:delta_full} the paired difference between HyInv and each baseline with its interval, and Figures~\ref{fig:metrics_heat} and~\ref{fig:metrics_heat_structure} recovery, diversity, pLDDT, scRMSD and lDDT.
Of the \BaseNCells{} comparisons (ten baselines in seven panel--predictor columns), all \BaseNPos{} point estimates favor HyInv and \BaseNSig{} have an interval that excludes zero; the other \BaseNZero{} include zero (\BaseZeroList).
PiFold, CMLM, LM-Design and DPLM-2 3B are single points at much lower diversity than HyInv, so their differences are not at matched diversity.

\begin{table}[!htbp]
\caption{\textbf{scTM of every method.} ESMFold2 on the five panels and RecentPDB under Boltz2 and Protenix2.}
\label{tab:full_sctm}
\centering
\footnotesize
\setlength{\tabcolsep}{4.2pt}
\begin{tabular}{@{}lrrrrr@{\hspace{10pt}}rr@{}}
\toprule
 & \multicolumn{5}{c}{ESMFold2} & \multicolumn{2}{c}{RecentPDB} \\
\cmidrule(lr){2-6}\cmidrule(lr){7-8}
Method & CATH4.3 test & CATH4.2 test & TS500 & TS50 & RecentPDB & Boltz2 & Protenix2 \\
\midrule
\rowcolor[HTML]{DCE6FA}
\textbf{HyInv} & \textbf{0.8399} & \textbf{0.8553} & \textbf{0.8987} & \textbf{0.9203} & \textbf{0.8605} & \textbf{0.8738} & \textbf{0.8781} \\
ProteinZero V2 (ESM2) & 0.8329 & 0.8473 & 0.8949 & 0.9125 & 0.8495 & 0.8634 & 0.8675 \\
ProteinZero V2 (ESM1) & 0.8328 & 0.8477 & 0.8948 & 0.9051 & 0.8496 & 0.8626 & 0.8664 \\
ProteinMPNN grouped & 0.8267 & 0.8437 & 0.8944 & 0.9158 & 0.8427 & 0.8602 & 0.8637 \\
ProteinMPNN & 0.8262 & 0.8443 & 0.8951 & 0.9065 & 0.8421 & 0.8615 & 0.8621 \\
LigandMPNN & 0.8233 & 0.8406 & 0.8893 & 0.9116 & 0.8290 & 0.8537 & 0.8555 \\
ProteinMPNN (low-noise, grouped) & 0.8239 & 0.8374 & 0.8929 & 0.9019 & 0.8301 & 0.8578 & 0.8621 \\
PiFold & 0.8182 & 0.8168 & 0.8837 & 0.8822 & 0.7937 & 0.8243 & 0.8374 \\
LM-Design 650M & 0.8140 & 0.8191 & 0.8806 & 0.8615 & 0.7840 & 0.8242 & 0.8322 \\
CMLM & 0.8093 & 0.8063 & 0.8794 & 0.8709 & 0.7741 & 0.8217 & 0.8303 \\
DPLM-2 3B & 0.7953 & 0.8093 & 0.8685 & 0.8696 & 0.7650 & 0.7862 & 0.7998 \\
\midrule
Native & 0.8465 & 0.8531 & 0.8959 & 0.8978 & 0.8231 & 0.8358 & 0.8419 \\
\bottomrule
\end{tabular}
\tablenote{Mean TM-score of the predicted structure to the native backbone, averaged over the designs of a target and then over targets; bold is the highest of the column among the designed rows (a point estimate: see Table~\ref{tab:delta_full} for the paired differences). Native is the refolded native sequence. The Boltz2 and Protenix2 columns fold the same RecentPDB sequences.}

\end{table}

\begin{table}[!htbp]
\caption{\textbf{HyInv minus baseline, scTM.} Difference in units of $10^{-3}$ with the paired 95\% interval in brackets; bold: the interval excludes zero.}
\label{tab:delta_full}
\centering
\footnotesize
\setlength{\tabcolsep}{5pt}
\begin{tabular}{@{}lccc@{}}
\toprule
Baseline & CATH4.3 test & CATH4.2 test & TS500 \\
\midrule
ProteinZero V2 (ESM2) & $+$7.0\,{\footnotesize\color{hymuted}[$-$0.6, 14.3]} & \textbf{$+$8.0}\,{\footnotesize\color{hymuted}[0.2, 16.3]} & \textbf{$+$3.9}\,{\footnotesize\color{hymuted}[0.0, 7.6]} \\
ProteinZero V2 (ESM1) & $+$7.0\,{\footnotesize\color{hymuted}[$-$1.3, 13.9]} & \textbf{$+$7.7}\,{\footnotesize\color{hymuted}[1.2, 13.9]} & $+$3.9\,{\footnotesize\color{hymuted}[$-$0.1, 7.8]} \\
ProteinMPNN grouped & \textbf{$+$13.2}\,{\footnotesize\color{hymuted}[2.5, 25.5]} & \textbf{$+$11.6}\,{\footnotesize\color{hymuted}[1.4, 25.9]} & \textbf{$+$4.3}\,{\footnotesize\color{hymuted}[0.0, 8.4]} \\
ProteinMPNN & \textbf{$+$13.6}\,{\footnotesize\color{hymuted}[2.3, 24.6]} & \textbf{$+$11.0}\,{\footnotesize\color{hymuted}[2.2, 22.5]} & $+$3.7\,{\footnotesize\color{hymuted}[$-$0.5, 7.8]} \\
LigandMPNN & \textbf{$+$16.6}\,{\footnotesize\color{hymuted}[8.1, 28.1]} & \textbf{$+$14.7}\,{\footnotesize\color{hymuted}[6.6, 26.1]} & \textbf{$+$9.4}\,{\footnotesize\color{hymuted}[4.4, 14.9]} \\
ProteinMPNN (low-noise, grouped) & \textbf{$+$16.0}\,{\footnotesize\color{hymuted}[4.3, 27.4]} & \textbf{$+$17.9}\,{\footnotesize\color{hymuted}[6.9, 34.1]} & \textbf{$+$5.9}\,{\footnotesize\color{hymuted}[1.4, 10.2]} \\
PiFold & \textbf{$+$21.6}\,{\footnotesize\color{hymuted}[6.1, 35.4]} & \textbf{$+$38.5}\,{\footnotesize\color{hymuted}[26.0, 56.6]} & \textbf{$+$15.0}\,{\footnotesize\color{hymuted}[9.5, 20.7]} \\
LM-Design 650M & \textbf{$+$25.9}\,{\footnotesize\color{hymuted}[11.1, 41.1]} & \textbf{$+$36.2}\,{\footnotesize\color{hymuted}[23.6, 54.9]} & \textbf{$+$18.2}\,{\footnotesize\color{hymuted}[12.0, 24.5]} \\
CMLM & \textbf{$+$30.6}\,{\footnotesize\color{hymuted}[14.2, 44.9]} & \textbf{$+$49.0}\,{\footnotesize\color{hymuted}[35.8, 67.8]} & \textbf{$+$19.4}\,{\footnotesize\color{hymuted}[14.5, 24.5]} \\
DPLM-2 3B & \textbf{$+$44.5}\,{\footnotesize\color{hymuted}[31.7, 55.2]} & \textbf{$+$46.0}\,{\footnotesize\color{hymuted}[34.7, 61.3]} & \textbf{$+$30.2}\,{\footnotesize\color{hymuted}[23.7, 36.7]} \\
\midrule
Baseline & TS50 & RecentPDB, ESMFold2 & RecentPDB, Boltz2 \\
\midrule
ProteinZero V2 (ESM2) & $+$7.8\,{\footnotesize\color{hymuted}[$-$7.0, 25.7]} & \textbf{$+$11.0}\,{\footnotesize\color{hymuted}[2.0, 20.6]} & \textbf{$+$10.4}\,{\footnotesize\color{hymuted}[1.7, 20.0]} \\
ProteinZero V2 (ESM1) & $+$15.2\,{\footnotesize\color{hymuted}[$-$1.5, 36.1]} & \textbf{$+$10.9}\,{\footnotesize\color{hymuted}[2.0, 20.0]} & \textbf{$+$11.2}\,{\footnotesize\color{hymuted}[3.5, 19.3]} \\
ProteinMPNN grouped & $+$4.5\,{\footnotesize\color{hymuted}[$-$17.9, 26.7]} & \textbf{$+$17.8}\,{\footnotesize\color{hymuted}[4.7, 32.5]} & \textbf{$+$13.6}\,{\footnotesize\color{hymuted}[3.5, 25.1]} \\
ProteinMPNN & $+$13.8\,{\footnotesize\color{hymuted}[$-$6.6, 37.0]} & \textbf{$+$18.5}\,{\footnotesize\color{hymuted}[7.3, 30.1]} & \textbf{$+$12.3}\,{\footnotesize\color{hymuted}[3.8, 21.6]} \\
LigandMPNN & $+$8.7\,{\footnotesize\color{hymuted}[$-$5.2, 23.5]} & \textbf{$+$31.6}\,{\footnotesize\color{hymuted}[18.2, 46.2]} & \textbf{$+$20.1}\,{\footnotesize\color{hymuted}[8.4, 33.0]} \\
ProteinMPNN (low-noise, grouped) & \textbf{$+$18.4}\,{\footnotesize\color{hymuted}[2.1, 37.3]} & \textbf{$+$30.4}\,{\footnotesize\color{hymuted}[16.0, 46.0]} & \textbf{$+$16.0}\,{\footnotesize\color{hymuted}[5.8, 27.4]} \\
PiFold & \textbf{$+$38.1}\,{\footnotesize\color{hymuted}[15.3, 65.5]} & \textbf{$+$66.9}\,{\footnotesize\color{hymuted}[43.8, 90.9]} & \textbf{$+$49.5}\,{\footnotesize\color{hymuted}[34.6, 66.6]} \\
LM-Design 650M & \textbf{$+$58.8}\,{\footnotesize\color{hymuted}[31.3, 90.2]} & \textbf{$+$76.5}\,{\footnotesize\color{hymuted}[55.3, 99.8]} & \textbf{$+$49.6}\,{\footnotesize\color{hymuted}[34.4, 65.7]} \\
CMLM & \textbf{$+$49.4}\,{\footnotesize\color{hymuted}[21.8, 82.9]} & \textbf{$+$86.4}\,{\footnotesize\color{hymuted}[63.5, 111.7]} & \textbf{$+$52.1}\,{\footnotesize\color{hymuted}[36.8, 69.1]} \\
DPLM-2 3B & \textbf{$+$50.7}\,{\footnotesize\color{hymuted}[18.4, 88.2]} & \textbf{$+$95.6}\,{\footnotesize\color{hymuted}[71.3, 121.5]} & \textbf{$+$87.6}\,{\footnotesize\color{hymuted}[66.8, 110.1]} \\
\midrule
Baseline & RecentPDB, Protenix2 &  &  \\
\midrule
ProteinZero V2 (ESM2) & \textbf{$+$10.5}\,{\footnotesize\color{hymuted}[2.0, 19.7]} &  &  \\
ProteinZero V2 (ESM1) & \textbf{$+$11.7}\,{\footnotesize\color{hymuted}[1.8, 21.9]} &  &  \\
ProteinMPNN grouped & \textbf{$+$14.4}\,{\footnotesize\color{hymuted}[3.8, 26.3]} &  &  \\
ProteinMPNN & \textbf{$+$16.0}\,{\footnotesize\color{hymuted}[5.2, 28.0]} &  &  \\
LigandMPNN & \textbf{$+$22.6}\,{\footnotesize\color{hymuted}[8.7, 37.1]} &  &  \\
ProteinMPNN (low-noise, grouped) & \textbf{$+$16.0}\,{\footnotesize\color{hymuted}[5.1, 28.4]} &  &  \\
PiFold & \textbf{$+$40.7}\,{\footnotesize\color{hymuted}[26.4, 56.3]} &  &  \\
LM-Design 650M & \textbf{$+$45.9}\,{\footnotesize\color{hymuted}[30.7, 62.9]} &  &  \\
CMLM & \textbf{$+$47.8}\,{\footnotesize\color{hymuted}[33.0, 63.8]} &  &  \\
DPLM-2 3B & \textbf{$+$78.3}\,{\footnotesize\color{hymuted}[57.9, 100.8]} &  &  \\
\bottomrule
\end{tabular}
\tablenote{HyInv minus baseline in units of $10^{-3}$ scTM, with the paired 95\% interval in brackets (CATH-topology cluster bootstrap; one cluster per target for TS500, TS50 and RecentPDB). Bold: the interval excludes zero.}

\end{table}

\subsection{scTM against diversity}
\begin{figure}[t]
\centering\includegraphics[width=\linewidth]{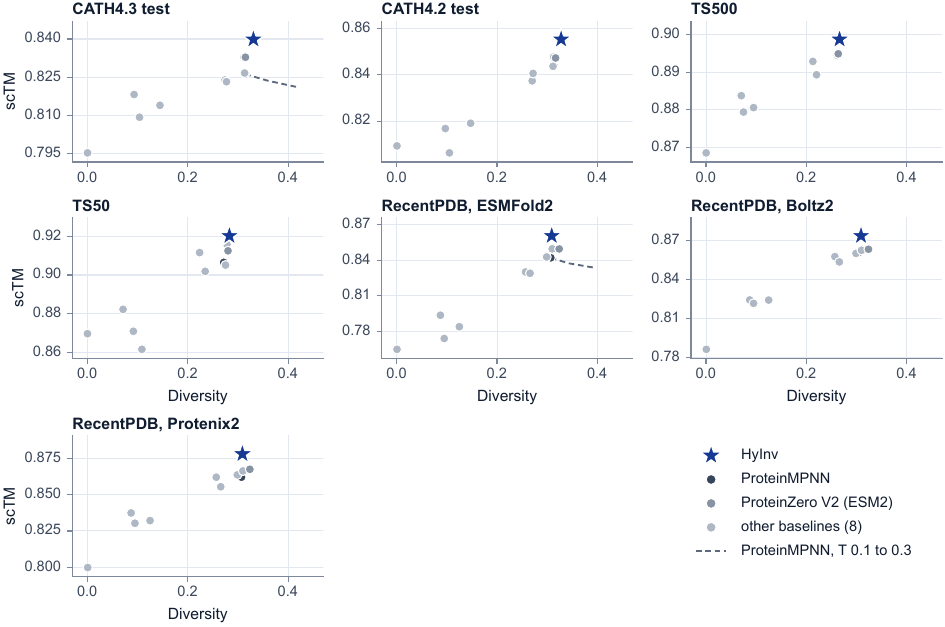}
\caption{\textbf{scTM against diversity on every panel.}
Gray: published baselines (darker: ProteinMPNN and ProteinZero V2); dashed: ProteinMPNN at three temperatures; star: HyInv.}
\label{fig:frontier_app}
\end{figure}

Figure~\ref{fig:frontier_app} places every method in the scTM--diversity plane of each panel.
ProteinMPNN is the only baseline with a temperature sweep (three temperatures; two were generated for this report).
Reading the quadratic fit through its three points at the HyInv diversity, HyInv is above the ProteinMPNN curve by \BaseFrontDeltaCathFour{} on CATH4.3 test and \BaseFrontDeltaCplt{} on RecentPDB (95\% intervals over clusters).

\section{Sampler details}\label{app:sampler}

\lstdefinelanguage{hysampler}{morekeywords={for,in,return},morecomment=[l]{\#}}
\begingroup\renewcommand{\lstlistingname}{Algorithm}
\begin{lstlisting}[float=!htbp,captionpos=t,language=hysampler,mathescape=true,basicstyle=\ttfamily\scriptsize,aboveskip=4pt,belowskip=4pt,
  caption={\textbf{Draft-and-rewrite sampling with fixed groups.} The draft temperature $\Tdraft$ and the rewrite temperature
  $\Trewrite$ are independent; $\Tdraft\to0$ replaces the draft by the argmax. Top-$p$ and min-$p$ optionally truncate each softmax.},label={alg:rewrite}]
Input: backbone $\bb$ with $L$ positions, model, seed $s$, groups $\Rrounds$, temperatures $\Tdraft$, $\Trewrite$
$c \leftarrow$ Encode($\bb$)                                     # once per backbone
$\sigma \leftarrow$ RandomPermutation(hash($s$, id($\bb$)), $L$)       # independent of structure and weights
$G_r \leftarrow \{\sigma_k : \lfloor k\Rrounds/L \rfloor = r\}$, $r = 0,\dots,\Rrounds-1$  # equal-size groups
$a \leftarrow$ (MASK, ..., MASK)
for $r = 0,\dots,\Rrounds-1$:                                   # blind draft
    $u \leftarrow$ Decode($c$, $a$)
    for $i \in G_r$: $a_i \sim \mathrm{softmax}(u_i/\Tdraft)$
for $r = 0,\dots,\Rrounds-1$:                                   # rewrite
    $\tilde a \leftarrow a$ with $\tilde a_i =$ MASK for $i \in G_r$
    $u \leftarrow$ Decode($c$, $\tilde a$)
    for $i \in G_r$: $a_i \sim \mathrm{softmax}(u_i/\Trewrite)$
return $a$
\end{lstlisting}
\endgroup

\paragraph{Procedure and groups.}
Algorithm~\ref{alg:rewrite} gives the procedure for one backbone.
The encoder runs once; the decoder runs on all $L$ sites in each of the $2\Rrounds=\DecodePasses$ passes, although only the sites of one group need new logits in a pass.
Without fixed residues, the draft starts from the all-masked state, so no sequence information enters before the first token is drawn.
A generator seeded by a SHA-256 hash of the seed and the chain identifier draws a uniformly random permutation of the sites, and the $k$-th site of the permutation receives group $\lfloor kR/L\rfloor$: groups are equal in size up to one site, independent of structure, sequence and weights, and unchanged by padding or batching; draft round $r$ and rewrite round $r$ use the same group.
The decoder accepts a partially masked state: each masked position is predicted using the backbone and the observed amino acid identities as sequence context.
For fixed groups the log-probability of a trajectory is the sum of the per-site conditionals of the draft and the rewrite rounds, $\log P(\tau)=\sum_{r}\sum_{i\in G_r}\big(\log p^{\mathrm d}_r(a^{\mathrm d}_i)+\log p^{\mathrm r}_r(a_i)\big)$, because the schedule does not depend on sampled tokens.
RL does not use this sampler: its rollouts are serial and its replay is one causal pass (Appendix~\ref{app:training}).

\paragraph{Random streams, truncation and fixed residues.}
Tokens are drawn by multinomial draws from the tempered distribution, with a separate stream for the draft; the seed of draw $k$ hashes the base seed, the chain keys of the batch and $k$, so exact sequences are reproducible for a fixed batch layout and the distribution is unaffected.
With $p=\operatorname{softmax}(z/T)$, min-p keeps tokens with $p\ge m\,p_{\max}$ and top-p the smallest set of most probable tokens with mass at least $q$ (the top token is always kept; both masks are computed on the same unrenormalized $p$ and intersected); truncation applies at every sampled stage, and the evaluated values are $q=\TopPValue$ and $m=\MinPValue$.
Pinned residues are handled by a wrapper around the network: it keeps their amino acid identities unmasked from the first draft round and redraws those same identities at every stage by replacing their logits with a one-hot distribution; pinned sites remain members of their groups.
A second sampler, used only to monitor RL on the development panel, commits the highest-probability proposals over eight steps; its diversity is lower than that of the rewrite sampler at the same nominal temperature (\ConfDivSameT{} against \RewriteDivSameT{} at $T=0.5$, supervised checkpoint), so temperatures of the two samplers are never compared.

\paragraph{Confirmation of the decoupled draft.}
Table~\ref{tab:sampler_gain} gives the gain over the unmodified sampler of the same checkpoint at equal diversity.
The settings were chosen on the development panel (Appendix~\ref{app:protocols}) and confirmed on CATH4.3 test and RecentPDB with new draws and fold seeds, on the checkpoint at update \RLPriorStep{}.
The gain of the decoupled draft at $\Tdraft=\TdraftVal$ is positive on all three panels with an interval that excludes zero (RecentPDB is the smallest panel and has the widest interval), and larger than that of top-$p$ or min-$p$ truncation.
With the unmodified sampler at the same development diversity, the final checkpoint differs from ProteinMPNN by \AblDeltaMPNNCathFour{} on CATH4.3 test and from ProteinZero V2 (ESM2) by \AblDeltaPZCathFour{} on CATH4.3 test and \AblDeltaPZCplt{} on RecentPDB (point estimates), so both the RL-trained checkpoint and the sampler contribute to the differences of Section~\ref{sec:results}.

\begin{table}[!htbp]
\caption{\textbf{Gain of the decoupled sampler over the unmodified sampler at equal diversity.}}
\label{tab:sampler_gain}
\centering
\footnotesize
\begin{tabular}{@{}lccc@{}}
\toprule
Setting & Development & CATH4.3 test & RecentPDB \\
\midrule
\rowcolor[HTML]{DCE6FA} \textbf{Decoupled draft, $\Tdraft=\TdraftVal$ (HyInv default)} & $+$0.0072\,{\footnotesize\color{hymuted}[$+$0.0062, $+$0.0118]} & $+$0.0078\,{\footnotesize\color{hymuted}[$+$0.0064, $+$0.0107]} & $+$0.0131\,{\footnotesize\color{hymuted}[$+$0.0065, $+$0.0196]} \\
Decoupled draft, $\Tdraft=\TdraftAltVal$ & $+$0.0054\,{\footnotesize\color{hymuted}[$+$0.0022, $+$0.0067]} & $+$0.0074\,{\footnotesize\color{hymuted}[$+$0.0060, $+$0.0096]} & $+$0.0089\,{\footnotesize\color{hymuted}[$+$0.0036, $+$0.0145]} \\
Top-$p$ \TopPValue & $+$0.0044\,{\footnotesize\color{hymuted}[$+$0.0028, $+$0.0063]} & $+$0.0032\,{\footnotesize\color{hymuted}[$+$0.0020, $+$0.0051]} & $+$0.0066\,{\footnotesize\color{hymuted}[$+$0.0018, $+$0.0119]} \\
Min-$p$ \MinPValue & $+$0.0033\,{\footnotesize\color{hymuted}[$+$0.0023, $+$0.0076]} & $+$0.0030\,{\footnotesize\color{hymuted}[$+$0.0017, $+$0.0050]} & $+$0.0068\,{\footnotesize\color{hymuted}[$+$0.0010, $+$0.0137]} \\
\bottomrule
\end{tabular}
\tablenote{Gain in scTM at fixed diversity over the unmodified sampler of the same checkpoint (difference of S@$D$, paired; $D=\SadRefDev$ on the development panel, \SadRefCathFour{} on CATH4.3 test and \SadRefCplt{} on RecentPDB; 95\% interval over CATH topology clusters, over chains on RecentPDB). Development chains $K=\KRef$, held-out panels $K=\KHeld$ with new draws and fold seeds; checkpoint at update \RLPriorStep.}

\end{table}

\section{Efficiency details}\label{app:efficiency}

\paragraph{Timing protocol.}
All latencies were measured on one \EffGPU{} of a node on which nothing else ran (PyTorch \ApxEffTorch, CUDA \ApxEffCuda, TF32 for matrix products off unless stated).
The timed region starts at host arrays and ends at the sampled tokens on the device: it contains the upload, the backbone batch, the encoder and structure terms and all \EffPasses{} decoder passes of every candidate batch, but not model loading, graph capture or kernel compilation (first call \EffFirstCallLo{} to \EffFirstCallHi{} per shape bucket).
Every call draws fresh seeds and nothing is cached across calls.
A cell is the median of \EffRepsHeadline{} repeats (headline cells), \EffRepsSweep{} (sweeps) or \EffRepsLib{} (slow systems of the library cells) after \EffWarmup{} warm-up calls; the interquartile range is below 1\% of the median in every ladder row, and repeating the headline cells in five fresh processes moves the median by at most \EffStabRange.
All systems design \EffNDesigns{} candidates per request with a maximum candidate batch of \EffBatch{}; ProteinMPNN and ProteinZero V2 use their published temperature, to which their cost is nearly insensitive.
Timing used a rewrite temperature of \EffTimingTr{} and the weights of the supervised stage, which have the same architecture and tensor shapes as the final checkpoint; the cost of the sampler does not depend on the values of the weights, supported by near-equal numerical differences of the BF16 path on the two sets (mean divergence \ApxEffKLSFT{} against \ApxEffKLMean).
The equivalence tests and the quality read-out used the RL checkpoint at update \EffQualityStep{} and were not repeated at update \RLStep{}; all benchmark sequences of Section~\ref{sec:results} come from the reference sampler.

\paragraph{The two numerical paths.}
The \emph{BF16 path} (HyInv in the main text) computes the edge messages and the node update with BF16 operands and FP32 accumulation, uses the hardware tanh form of GELU and keeps FP32 inputs and outputs; it is a numerical variant of the reference sampler.
Against the reference on \ApxEffNumBackbones{} development backbones the largest absolute logit difference is \ApxEffLogitMax, the mean Kullback--Leibler divergence per sampled position \ApxEffKLMean{} (99th percentile \ApxEffKLP) and the argmax agrees at \ApxEffArgmax{} of the positions; with the same seeds \EffAgreeSeqFull{} of the sequences are identical to the reference and \EffAgreeSiteFull{} of the positions agree, so its effect on quality is measured on the distribution of designs (below).
The \emph{exact FP32 path} computes its matrix products with a three-pass TF32 emulation of FP32 and neither BF16 operands nor the tanh form of GELU.
Its sequences equal those of the reference for the same seeds (\EffExactIdentEager{} eager, \EffExactIdentGraph{} with graph capture and \EffExactIdentPadded{} in padded multi-target batches, including fixed positions, missing atoms, short chains and a synthetic two-chain input), and graph replay equals eager execution bit for bit; only this path is called identical.

\begin{figure}[t]
\centering\includegraphics[width=\linewidth]{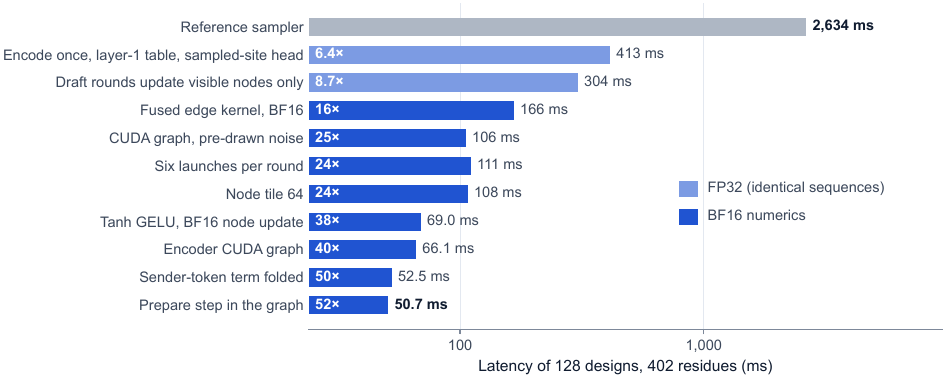}
\caption{\textbf{All eleven optimization steps.} Latency of \EffNDesigns{} designs for the \EffLenLong-residue backbone after each cumulative step; light blue: exact FP32 path (identical sequences), blue: BF16 numerics. The draft node subset includes positions with observed amino acid identities and positions sampled in the current round. Two steps are not faster than their predecessors and are kept for completeness.}
\label{fig:eff_ladder_full}
\end{figure}

\paragraph{Optimizations.}
Figure~\ref{fig:eff_ladder_full} adds the eleven items one at a time; later items depend on earlier ones, so removing one item at a time from the final configuration is the better estimate of its contribution.
Five groups of items carry most of the gain.
\emph{Shared work:} all candidates of a request share one backbone, so the encoder, the structure terms and the first draft round (with all non-fixed sequence positions initially masked) are computed once, and the last decoder layer and output head run at the sampled positions only.
\emph{Restrict draft updates to context and active positions, and apply one linear map after the neighbor sum:} the node subset includes positions with observed amino acid identities and those sampled in the current round; removing the node subset, the shared round or the late map slows the exact path by \ApxEffLooSubset{}, \ApxEffLooShare{} and \ApxEffLooW{} on the \EffLenLong-residue request.
\emph{Fused edge kernel:} gather, GELU, the second edge product and the neighbor sum form one Triton kernel, so edge activations are never written out; its gain comes with BF16 operands.
\emph{Pre-drawn noise and CUDA graphs:} each round is captured in one graph per shape bucket, which matters most for short backbones (this item also enables TF32 for the remaining linear layers).
\emph{Fewer launches and numerics:} six kernel launches per round, tanh GELU (erf everywhere makes the request \EffLooAllGeluLong{} slower) and a folded token term (removal costs \EffLooFoldLong).
The last rung of the ladder (\EffAblationEnd) and the headline latency (\EffLatencyLong) are two separately timed runs of the same configuration.

\paragraph{Baseline timings.}
The baselines of Figure~\ref{fig:eff_bars}a run with their released implementations (FP32 without graph capture or compilation; DPLM-2 3B in BF16 autocast), each at its published setting.
Where the released adapter handles one structure per call, \EffNDesigns{} designs are \EffNDesigns{} calls, so a batched re-implementation would be faster; because DPLM-2 3B decodes by argmax, every call returns the same sequence, and its time is that of producing \EffNDesigns{} designs scaled from eight timed calls.
The speed-ups in the main text are against these implementations, not against the best achievable speed of the baseline models.

\paragraph{Library scale and memory.}
A library of \EffLibN{} sequences for one backbone of \EffLibLen{} residues takes \EffLibTime{} on one GPU (\EffThroughput; reference sampler \EffLibRefTime, ProteinMPNN \EffLibMPNNTime), \EffLibEightGPUTime{} on eight GPUs (one process per GPU on disjoint parts of the library within one node, \EffThroughputEightGPU, \EffScaling{} times the single-GPU rate at the same batch size) and a single-GPU rate of \EffLibBatchRate{} at candidate batch \EffLibBigBatch.
For \EffNDesigns{} candidates at \EffLenLong{} residues and a batch of \EffBatch{} the peak memory allocated by tensors is \EffMemoryMiB{} for HyInv (BF16), \EffMemoryRefMiB{} for the reference sampler and \EffMemoryMPNNMiB{} for ProteinMPNN.

\paragraph{Quality of the BF16 path.}
Its quality is judged on the distribution of designs, not on single sequences: the quantity is the S@D of the development read-out (scTM at diversity \ApxEffDZero{} from a quadratic fit through four rewrite temperatures, ESMFold2, \KRef{} designs per chain, draft temperature \TdraftVal) on the RL checkpoint at update \EffQualityStep, and Table~\ref{tab:eff_quality} gives the difference (BF16 path minus reference sampler), paired by chain, on the whole development panel of \EffNDevTargets{} chains and on the three disjoint subsets it consists of.
On the whole panel the difference is \EffSadDelta{} with interval \EffSadInterval, whose lower end is above the margin of $-$\ApxEffCritical{} fixed for this check.
Subsets of about five hundred chains cannot resolve a difference of that size (interval half-widths of two to three thousandths, point estimates on both sides of zero: \EffSadASub, \EffSadBSub, \EffSadCSub).
Per rewrite temperature the scTM difference has mixed signs without a trend, and the diversity of the two samplers agrees to within \ApxEffDivAgree.
The check covers the development panel, ESMFold2, \KRef{} designs and rewrite temperatures up to \ApxEffTmax; test panels, the independent predictors and other checkpoints are not covered by it.
Shorter schedules cost more quality than they save time: the S@D of six schedules with fewer rounds (6+8, 4+8, 8+6, 8+4, 6+6 and 4+4 draft and rewrite rounds) is lower than that of the \EffPasses-pass default by \EffRoundsMin{} to \EffRoundsMax, so the default of \DraftRounds{} draft and \RewriteRounds{} rewrite rounds was kept.

\begin{table}[!htbp]
\caption{\textbf{BF16 path against the reference sampler on the development panel.}}
\label{tab:eff_quality}
\centering
\footnotesize
\begin{tabular}{@{}lrrcc@{}}
\toprule
Chains & Targets & Clusters & S@D difference & Lower end $\geq-0.002$ \\
\midrule
Subset 1 & 512 & 51 & $-$0.0008\,{\footnotesize\color{hymuted}[$-$0.0021, $+$0.0022]} & no \\
Subset 2 & 512 & 50 & $+$0.0005\,{\footnotesize\color{hymuted}[$-$0.0013, $+$0.0032]} & yes \\
Subset 3 & 505 & 61 & $+$0.0001\,{\footnotesize\color{hymuted}[$-$0.0032, $+$0.0018]} & no \\
\midrule
\rowcolor[HTML]{DCE6FA} \textbf{Whole development panel} & 1\,529 & 91 & $-$0.0001\,{\footnotesize\color{hymuted}[$-$0.0009, $+$0.0016]} & yes \\
\bottomrule
\end{tabular}
\tablenote{S@D difference, BF16 path minus reference sampler, at the development diversity $D=\ApxEffDZero$ (four-point quadratic fit, paired, 95\% interval over CATH topology clusters). The whole panel is the union of the three disjoint subsets.}

\end{table}

\section{Application protocols}\label{app:applications}
The applications of Section~\ref{sec:applications} use the HyInv default setting (the final checkpoint at update \RLStep, draft temperature \TdraftVal, rewrite temperature \TrewriteVal) without task-specific tuning.
They report predicted-structure agreement, local geometry, sequence statistics, diversity and novelty; none is an experimental measurement.
Panel rules, thresholds and endpoints were fixed before designs were generated, and panels were chosen from native sequences and structures only.

\paragraph{De novo backbones.}
\AppxDenovoN{} de novo backbones in \AppxDenovoClusters{} clusters of 30\% sequence identity, none with a hit that passes the homology rule below, with \KHeld{} designs per backbone; the control is a set of \AppxNativeControlN{} length-matched native chains.
A design is \emph{foldable} if its ESMFold2 scTM is at least 0.9 and \emph{novel} if its sequence identity to the nearest UniRef50 sequence and to the nearest sequence of the pretraining library is below 0.3; the yield is the number of novel-and-foldable designs per backbone.
Both readings are reported, fixed before designs were generated: the yield of all designs, and the yield of designs that pass a composition filter (sequence entropy at least \DemoFiltEntropy~bits, hydrophobic fraction (AVILMFWY) between \DemoFiltHydLo{} and \DemoFiltHydHi, longest homopolymer run at most \DemoFiltHomopolymer), which excludes low-complexity sequences that a structure predictor can fold to a high scTM.
The filter lowers the yield of ProteinMPNN and ProteinZero V2 more than that of HyInv; the unfiltered yield of ProteinMPNN is \DemoDenoMPNNGridLo{} to \DemoDenoMPNNGridHi{} over its three lowest temperatures, which span the diversity of HyInv, and \HyInv\ is above both baselines in both readings.
HyInv is compared with ProteinMPNN and ProteinZero V2 over their temperature grids, interpolating each curve at the HyInv diversity without extrapolating by more than 0.05; the resampling unit is the cluster of 30\% identity.

\paragraph{Pocket-constrained libraries.}
\AppxLibN{} ligand-pocket backbones: ten chosen by rule (length range, complete backbone atoms, a non-solvent ligand with at least eight residues within 4~{\AA}, no hit that passes the homology rule), in hash order, and two single cases chosen earlier.
Residues within 4~{\AA} of the ligand keep their native identity (the pinned-position wrapper of the sampler for HyInv, a mask on the free positions for the baselines, so retention is a check of the code and not a result), and \AppxLibPerCell{} designs are made per backbone and cell.
A design passes if its scTM is at least \AppxLibScTMThresh{} and its site C$\alpha$ RMSD at most \AppxLibSiteThresh{}~{\AA}; the endpoint is the number of different sequences, after clustering at 50\% identity, among the passing designs of a plate of \AppxLibPlateSmall{} or \AppxLibPlateLarge{} designs (\AppxLibPlates{} random plates per cell), against ProteinMPNN at the temperature whose diversity at the free positions is closest to that of HyInv.
Controls are the native refold, a random substitution with the same number of mutations and HyInv with nothing pinned.

\paragraph{Loop and window redesign and named enzymes.}
One loop (the longest coil run) or one window of 30\% of the chain is redesigned with the rest of the sequence fixed, on \AppxRegionNStrict{} backbones chosen by rule and on \AppxRegionNNamed{} named enzymes (TEM-1 $\beta$-lactamase, subtilisin, lysozyme, PETase) whose catalytic residues are fixed too; \AppxRegionPerCell{} designs per backbone and region, \AppxRegionDesigns{} in total.
A design succeeds if its whole-chain scTM is at least 0.9 and the C$\alpha$ RMSD of the redesigned region after fitting the unchanged context is at most 1.5~{\AA}; the catalytic-site RMSD is the heavy-atom RMSD of the pinned catalytic residues after the same fit.
The success rate is read only where the native refold meets the criterion on a majority of backbones.
Comparators are ProteinMPNN at the temperature matched to the diversity of the redesigned region, ProteinZero V2 and the native refold.
The named enzymes are homologous to the training data and chosen for display: they illustrate the preservation of a fixed geometry, not the design of new enzymes.
Figure~\ref{fig:demo_enzyme} reports the catalytic-site RMSD of three of them (TEM-1 $\beta$-lactamase, subtilisin and PETase).

\paragraph{Homology of the panels.}
The native chain of each panel backbone was searched against the pretraining library and the benchmark reference set (MMseqs2); a hit counts if its identity is at least 0.3 and it covers at least 0.8 of the query or of the target.
This is a coverage rule: hits that cover less of the chain remain, so a backbone passing the rule is homology-filtered at these thresholds, which is weaker than having no similar fragment in the library (Table~\ref{tab:app_homology}).
Strict backbones have no passing hit, relaxed backbones relax one criterion of the strict rule (fixed before any design), and the named enzymes are a fixed list; the table documents the panel rule only.

\begin{table}[!htbp]
\caption{\textbf{Homology screen of the application panels.}}
\label{tab:app_homology}
\centering
\footnotesize
\begin{tabular}{@{}lrrr@{}}
\toprule
Panel & Backbones & Passing hit & Partial hits only \\
\midrule
De novo backbones & 74 & 0 & 28 \\
Pocket library & 12 & 0 & 7 \\
Loop and window, strict & 2 & 0 & 2 \\
Loop and window, relaxed & 6 & 0 & 4 \\
Named enzymes & 4 & 4 & 0 \\
\bottomrule
\end{tabular}
\tablenote{A hit passes the rule if its sequence identity to the pretraining library or to the benchmark reference set is at least 0.3 and it covers at least 0.8 of the query or of the target; partial hits stay below the coverage limit. Strict: no passing hit; relaxed: one criterion relaxed before any design was made. The named enzymes are homologous to training data by design.}

\end{table}

\paragraph{Predictors, comparators and statistics.}
Monomers are folded with ESMFold2 in the recipe of Appendix~\ref{app:protocols} with one fold seed shared by all generators for the same backbone and chunk, so that generators are paired; ESMFold2 is also the reward predictor, so a subset of cells is folded with Boltz2 and Protenix2 as well.
Comparators are ProteinMPNN and ProteinZero V2 only (the latter is declared for chains of up to 100 residues, which is stated where it is used on longer chains).
The temperature of the diversity-matched baseline is chosen from sequences alone, before any fold is read, as the grid value whose mean pairwise Hamming distance over the designed positions is closest to that of the HyInv default; where HyInv has lower diversity than the lowest baseline temperature the match is as close as the grid allows.
Differences are paired by backbone, averaged over designs first, with 95\% percentile intervals from \BootB{} resamples (with four to twelve backbones the intervals are wide).
Intervals are not adjusted for multiple comparisons.

\section{Release and reproducibility}
\paragraph{Release.} We will release the \HyInv\ model weights and inference code at \url{https://github.com/Tencent-Hunyuan/hyra-results/tree/main}.
\paragraph{Reproducibility.} Every number, table and figure in this report is generated by scripts from result files. Sampling is random, so a rerun with other seeds changes individual designs within the reported intervals.

\end{document}